\documentclass{article}

\usepackage{PRIMEarxiv}
\usepackage[utf8]{inputenc} % allow utf-8 input
\usepackage[T1]{fontenc}    % use 8-bit T1 fonts
\usepackage[hidelinks]{hyperref} % hyperlinks
\usepackage{url}            % simple URL typesetting
\usepackage{booktabs}       % professional-quality tables
\usepackage{amsfonts}       % blackboard math symbols
\usepackage{nicefrac}       % compact symbols for 1/2, etc.
\usepackage{microtype}      % microtypography
\usepackage{fancyhdr}       % header
\usepackage{graphicx}       % graphics
\graphicspath{{media/}}     % organize your images and other figures under media/ folder
\usepackage{amsmath}
\usepackage{amssymb}
\usepackage{tabularx}   % in preamble
\usepackage{multirow}
\usepackage{algorithm}
\usepackage{algpseudocode}
\usepackage{fourier} 
\usepackage{array}
\usepackage{makecell}
\usepackage{commath}
\usepackage{subcaption}
\usepackage{xcolor}
\title{Robust Recurrent Reinforcement Learning under Evolving Hidden Disturbances with Application to Rover Wheel Slip
\thanks{The original methodological study was conducted while Saki Omi was affiliated with Cranfield University.}
}

\author{
Saki Omi$^{1,*}$,
Hyo-Sang Shin$^{2}$,
Namhoon Cho$^{3}$,
Antonios Tsourdos$^{2}$,
Miguel A. Olivares-Mendez$^{1}$
\\[1ex]
$^{1}$Space Robotics Research Group, Interdisciplinary Centre for Security, Reliability and Trust, University of Luxembourg
\\
$^{2}$School of Aerospace, Transport and Manufacturing,
Cranfield University, Cranfield, U.K.
\\
$^{3}$Department of Aerospace Engineering, Seoul National University, Seoul, South Korea
\\[1ex]
$^{*}$Corresponding author:
\texttt{saki.omi@uni.lu}
}

\begin{document}
\maketitle
\begin{abstract}\label{abstract}

Reinforcement learning (RL) has shown strong performance in continuous-control tasks, but its application to systems affected by evolving hidden disturbances remains challenging. Such problems are naturally formulated as partially observable Markov decision processes, where the agent must construct a decision-relevant internal representation from interaction history. Recurrent neural networks are commonly used for this purpose, yet the effects of information selection and recurrent architecture remain insufficiently understood. This study investigates how observation history, action history, history length, and network structure affect the robustness of recurrent Twin Delayed Deep Deterministic Policy Gradient (TD3) agents. Three recurrent architectures are introduced and evaluated under controlled disturbances with different temporal characteristics. The results show that action history is particularly important when observed system responses depend on previously applied actions, and that processing past and current action--observation information within a unified temporal sequence improves performance compared with processing them through separate branches. We also introduce H-TD3, which reuses recurrent states generated by the actor to initialize the critic and thereby reduces duplicated sequence processing. To assess the potential engineering benefit of the investigated architectures, they are further tested in a simulation-based differential-drive rover motion-regulation task under hidden asymmetric wheel slip. The recurrent architectures retain their advantage under the physically motivated multiplicative wheel-slip model, while policies trained with abstract temporally structured disturbances transfer more effectively to previously unseen wheel-slip dynamics than policies trained without disturbances. These findings provide practical guidance for designing recurrent RL agents that operate under partial observability and time-varying hidden disturbances.

\end{abstract}

\keywords{TD3 \and Recurrent neural network \and POMDP \and robustness \and causality}
\section{Introduction}\label{intro}

Machine learning (ML) techniques have
been widely applied to control problems because they can learn decision-making policies directly from data without requiring a complete analytical model of the system.
In particular, deep reinforcement learning (RL) has achieved
strong performance in continuous-control tasks across standardized benchmark environments
such as OpenAI Gym \cite{OpenAI-Gym}.

However, these successes often rely on the assumption of a Markov Decision Process (MDP),
in which the agent has access to a state representation that is sufficient for decision making.
In real-world applications, this assumption
is often violated.
Sensor limitations, noise, and unmodeled disturbances introduce partial observability, making the Partially Observable Markov Decision Process (POMDP) framework more suitable \cite{kurniawati2021partially,hernandez2019survey}.

To address partial observability, agents must infer
decision-relevant information about the latent state
from interaction histories. Recurrent neural networks (RNNs)—particularly Long Short-Term Memory (LSTM) and Gated Recurrent Unit (GRU) architectures—have been employed to model temporal dependencies and generate internal representations that
summarize past interactions
\cite{miki2022learning,li2015recurrent,hausknecht2015deep,singla2019memory}. These methods have demonstrated improved performance in POMDP settings by maintaining memory of past interactions.

Despite this progress, most studies focus
primarily on observation histories, while past actions are often omitted.
Since actions play a causal role in state transitions, their exclusion can limit the
informativeness
of the internal representation. The selection of information, including both observations and actions, is therefore
important for constructing internal state representations under partial observability.
Although some prior work highlights the importance of including action sequences \cite{zhu2017improving}, systematic investigation
across disturbances with different temporal characteristics and recurrent architectural designs
remains limited.

Alternative temporal modeling approaches, such as Transformer-based networks
\cite{janner2021offline,chen2021decision} and Liquid Time-Constant (LTC) models
\cite{hasani2021liquid}, have also been explored. However,
the influence of information selection and temporal structure remains
difficult to isolate when increasingly complex sequence models are used, and their
robustness across disturbances with different temporal characteristics is still
insufficiently understood.
Additionally, while much attention has been given to Sim2Real transfer,
many approaches emphasize increasing simulator fidelity
\cite{shen2021igibson,freeman2021brax}. In practice,
a simulator cannot reproduce every aspect of the deployment environment
\cite{truong2022rethinking}, underscoring the need to enhance the robustness and
generalization of RL agents
to unmodeled disturbances.

Another challenge lies in the computational cost of training RNN-based RL agents. In off-policy algorithms such as
Twin Delayed Deep Deterministic Policy Gradient (TD3)
\cite{fujimoto2018addressing},
the same interaction history may be processed separately by the recurrent actor and critic networks, increasing the computational cost as the history length grows. Although fully sharing the recurrent encoder can reduce this duplication, shared recurrent actor--critic architectures have been reported to hinder learning compared with architectures using separate encoders \cite{ni2021recurrent}. This trade-off motivates architectures that reduce redundant recurrent computation without requiring the actor and critic to share the entire recurrent representation.

This paper investigates how information selection, history length, and
recurrent network structure affect the robustness of RL agents operating under partial observability and evolving disturbances.
Controlled benchmark environments are first used to isolate the effects of observation history, action history, and architectural design under disturbances with different temporal characteristics. To assess the potential engineering benefit of the resulting architectures, we then test them in a simulation-based differential-drive rover motion-regulation task under hidden asymmetric wheel slip.
The main contributions are as follows:

\begin{enumerate}
    \item We investigate the effect of selecting observation and action histories, and show that incorporating action sequences
    improves performance under partial observability by providing information about how previous control inputs influenced the observed system response
    (Sections \ref{keyidea}--\ref{actionInclusion}).

    \item We propose
    recurrent TD3 architectures that process past and current action--observation information within a unified temporal sequence, enabling the recurrent representation to incorporate the current interaction consistently with the preceding history
    (Section \ref{structure}).

    \item A novel algorithm, H-TD3, is introduced, in which the critic reuses hidden states generated by the actor,
    reducing duplicated recurrent sequence processing between the actor and critic while retaining separate downstream networks
    (Section \ref{structure}).

    \item
    We evaluate the proposed architectures across controlled benchmark disturbances with different temporal characteristics and in a simulation-based, physically motivated differential-drive rover case under hidden asymmetric wheel slip. The rover study shows that the advantage of action-history-based recurrent architectures extends to multiplicative wheel-slip dynamics and that training with abstract temporally structured disturbances can improve zero-shot transfer to previously unseen wheel-slip dynamics
    (Sections \ref{discussion} and \ref{engineering2}).
\end{enumerate}

This work aims to support the development of RL algorithms that are
more robust to evolving hidden dynamics by connecting controlled architectural analysis with evaluation in a simulation-based, physically motivated robotic control task.

The remainder of this paper is organized as follows. Section~\ref{relatedWork} reviews related work, and Section~\ref{background} introduces the required background. Section~\ref{keyidea} discusses the role of action history in partially observable control, while Section~\ref{actionInclusion} investigates the effects of information selection and history length. Section~\ref{structure} presents the proposed recurrent TD3 architectures, including H-TD3. Section~\ref{discussion} evaluates the architectures under controlled benchmark disturbances. Finally, Section~\ref{engineering2} presents the differential-drive rover engineering case under hidden wheel slip, followed by the conclusion in Section~\ref{conclusion}.

\section{Related Work}\label{relatedWork}
\paragraph{RL in Dynamic Environments}
Recent advancements in RL are casting more interest in discussing the challenges posed by dynamic and partially observable environments toward real-world implementation. A comprehensive survey highlighting the challenges and strategies for RL in dynamic environments was provided in \cite{padakandla2021survey}. The survey emphasized the importance of the adaptability of the “active” context to perform in dynamically varying environments. The strategy to detect the context is discussed in \cite{zhang2023dynamics} so that the agent can identify changes in environmental dynamics. Also, \cite{martinez2022deep} presented a motion planner and a navigation algorithm for dynamic environments by incorporating a model called the Dynamic Objective Velocity Space model which reflects the dynamics of the scenario. In addition, new approaches, such as LTC networks \cite{hasani2021liquid} and decision transformers \cite{janner2021offline, chen2021decision}, to perform dynamic and partially observable tasks are demonstrating promising results in the robust performance of the agent by emphasizing sequential modeling and temporal dynamics. Even though various approaches are considered, utilizing RNN remains one of the most common and well-known approaches to cope with the dynamic environment in RL. Our study is focused on the method used in this aspect. The existing algorithms are often evaluated in several POMDP environments with different levels of complexity. However, this approach does not allow us to observe the detailed behavior which would depend on the characteristics of time-varying disturbance. Therefore, the algorithms in this work are evaluated in one simple environment with various types of disturbances.  
\paragraph{Action Sequence Inclusion}
The sequential information is processed by the RNN identifying the hidden dynamics of the environments. The early example of LSTM utilized in RL would be RL-LSTM \cite{harmon1996multi}. The implementation of LSTM was motivated to solve grid T-maze which required the agent to remember the action from the past. Only sequential observation was utilized, but RL-LSTM could effectively solve this non-Markovian scenario. Deep Recurrent Q-learning (DRQL) \cite{hausknecht2015deep} was developed more recently by replacing the first dense layer after a convolutional network of DQN \cite{mnih2015human}. DRQL demonstrated that the agent could perform in Atari games in POMDP conditions that DQN could not solve. However, the original DRQL also utilized only observation trajectory. Later, the importance of action sequence inclusion was pointed out as the action information was crucial for belief estimation \cite{zhu2017improving}. The comparisons between DRQN and the developed action-specific deep recurrent Q-learning Network (ADRQN) were represented based on simulation results. However, only one type of disturbance was tested in two different environments and it requires more study to understand the effectiveness of the inclusion of the action in different situations. The examples of the RL algorithm with RNN without utilizing sequential action in actors can be found in \cite{bae2023deep,everett2018motion} while they are required to perform in dynamic environments. Although both of the applications showed notable results, the consideration of the action sequence would further improve their performance. It is also a common approach to separate the estimation and control strategy using information states in RL for POMDP conditions \cite{malikopoulos2022separation,gasse2021causal}. In this approach, the information states are explicitly updated based on Bayesian inference. Then, the control strategies are learned by taking the information state into account. The suggested approach supports the idea of considering past action information. In many cases, the discussion of the selection of the information is overlooked. This study discusses the importance of the action history inclusion based on theory and also simulation experiments with extensive scenarios.
\paragraph{History Length}
Another aspect to be explored for RL with RNN is the sufficient length of the information. For instance, in DRQL development \cite{hausknecht2015deep}, two types of updates were considered; Bootstrapped Sequential Updates and Bootstrapped Random Updates. The former randomly selects episodes and selected full episodes are replayed from the first step to the end. Hence, the hidden state of LSTM is carried out without initialization throughout one episode. The latter case randomly selects the point of the episode and unrolls the experience for certain time steps. Bootstrapped Sequential Updates can capture longer sequential information but it violates the random sampling policy of DQN. Bootstrapped Random Updates can keep the random sampling but capture the information in a shorter time scale since it requires hidden states of LSTM to be initialized at the beginning of each step. Both types showed similar performance in their investigation, and Bootstrapped Random Updates was finally applied for further discussion because it takes a shorter time to train. In another example, Recurrent Deterministic Policy Gradient algorithm (RDPG) developed by Heess et al. \cite{heess2015memory} utilizes the update method similar to Bootstrapped Sequential Updates in DRQ. During the update, a minibatch of $N$ episodes is sampled and the target values are computed for each sampled episode through Back-Propagation Through Time (BPTT). Also, the performance of the model-free RL algorithms with histories of entire episodes was investigated by Yang et al. motivated by the absence of the investigation in the existing literature \cite{yang2021recurrent}. The study found challenges when solving a task that requires action exploration and provides sparse rewards. Furthermore, the different length was investigated in \cite{ni2021recurrent} and mentioned that the ideal length seemed to be problem-specific and may have required tuning. However, the discussion of the length of the sequence is not prioritized and this area is not explored enough. The history length compromises the performance and computational burden. Further investigation and tuning guidance are required and our work also delivers the results by changing the length of the history.  
\paragraph{Network Architecture}
Various RNN architectures are also found in the literature. For example, the observation and action are concatenated at the input layer in actor in \cite{nguyen2023equivariant} and the data is processed in one stream. In critic, the current action is added from the middle of the network stream after the history is processed by RNN at the critic. In another example, the network architecture can have feedforward and recurrent branches instead of processing data in one stream \cite{peng2018sim}. The network architecture found in \cite{bae2023deep} separated the stream into two after the entrance for feedforward and recurrent network layers. The current data is processed twice in both streams. It is rare to find explicit explanations and interpretations of the architecture in the literature. In our study, we work on the interpretation that follows the belief state estimation which would improve the robustness of the optimality in the POMDP condition.  
\paragraph{Computational Efficiency}
There are only a few discussions on computational efficiency in this context. In \cite{ni2021recurrent}, shared and separated recurrent actor-critic architectures are investigated. It was concluded that the shared architecture failed to learn some tasks while it has an advantage in computational efficiency to train the system. In our work, we introduce an algorithm, named H-TD3, that tackles the inefficiency of the RNN-based RL by sharing the hidden state of the actor network with the critic network. 

\section{Preliminaries}\label{background}

In \cite{meng2021memory}, the Long Short-Term Memory (LSTM) \cite{hochreiter1997long} layer was implemented in TD3 \cite{fujimoto2018addressing} algorithm which is named LSTM-TD3. LSTM-TD3 has LSTM layers in actor and critic networks. The past sequential information is processed through LSTM layer. Processing sequence in an entire episode is time-consuming to train the network. Hence, in LSTM-TD3, the networks process partial trajectory instead of entire trajectory up to the current time step by setting a specific sequence length $l$ to generate the action from the policy and to roll out the trajectory during the update as a hyper-parameter. This method to train a network with the partial history of a specific length is also applied in \cite{hausknecht2015deep} as described in Section \ref{relatedWork}.

\begin{figure}[h]
\begin{center}
\noindent
  \includegraphics[width=0.9\linewidth]{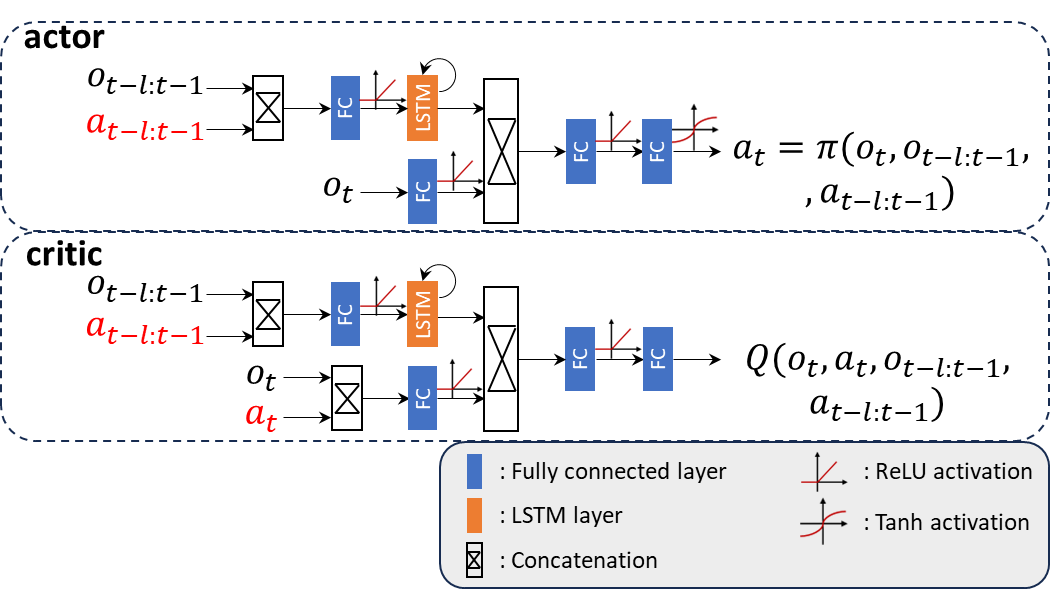}
  \caption{LSTM-TD3 schematics}\label{fig:lstmtd3}
\end{center}
\end{figure}
Another aspect to be mentioned is that LSTM-TD3 has a double input channel as network architecture as shown in Fig. \ref{fig:lstmtd3}. The input data for the first input channel is sequential past information $[o_{t-l:t-1}, a_{t-l:t-1}]$ from the memory and the second input channel takes the current information $[o_{t} (, a_{t})]$. The expressions $o_{t-l:t-1}$ and $a_{t-l:t-1}$ describe the sequential observation and action between time step $t-l$ and $t-1$ respectively. This structure was selected in order to prioritize the most recent observation since it is considered that the recent observation is more valuable in decision-making than earlier observations. Note that this hypothesis is questionable because it might not be true that the latest observation is more informative to reveal the true state of the system. This will be discussed in detail in Section~\ref{structure}. LSTM-TD3 \cite{meng2021memory} was tested for robot locomotion tasks in open AI gym simulation environment \cite{OpenAI-Gym} with several types of POMDP such as removing one of the state elements from the observation and adding random noise to observation. Especially in the scenario where one or all of the observation is lost with probability $p$, LSTM-TD3 showed greater performance compared to original TD3 and other Deep RL algorithms.

\section{Dynamic RL Adapting to Environment with Disturbance}\label{keyidea}
The challenge of POMDP in this study is that we have no access to the true state of the system $s_t$ due to the disturbances in observations. In our scenarios, disturbances are realized by stochastic or dynamic noise, or by hiding one of the elements of the observation. The details of disturbances are defined in Section \ref{actionInclusion}. If $s_t$ can be estimated or represented in an internally encoded format, it is possible to restore the MDP condition. Then, an already established RL algorithm can be utilized to identify the (sub-)optimal policy. Therefore, recovering the MDP condition through certain architecture design choices such as the definition of input/output variables should be important for the policy robustness. Hence, our focus is to develop an RL agent that can dynamically adapt to observations affected by disturbances with their own dynamics. To make a dynamic RL agent, the system should be able to generate the internal state representation by assessing causal information from the trajectory.

\subsection{Causality and Statistics}
The dichotomy of statistics and causality is emphasized in \cite{pearl2000models}. A statistical model works on the joint probability of observed variables. A causal model, on the other hand, assumes that certain variables are unobserved. The causal analysis enables inference of dynamics of events under changing conditions, not only the likelihood of events under static conditions. 

In MDP, a static transition model is available (i.e., $P(s_{t+1}\vert s_{t}, a_t)=P(o_{t+1}\vert o_{t}, a_t)$). However, in POMDP with dynamic disturbances, the model of the world dynamically changes while the true state is hidden from the observation. Therefore, solving POMDP requires an additional mechanism to identify the causal effect and recognize the pattern based on available information in the past and current. 

\subsection{Analysis of Disturbance}
\begin{figure*}[!t]
\centering
\subfloat[]{\includegraphics[width=0.45\textwidth]{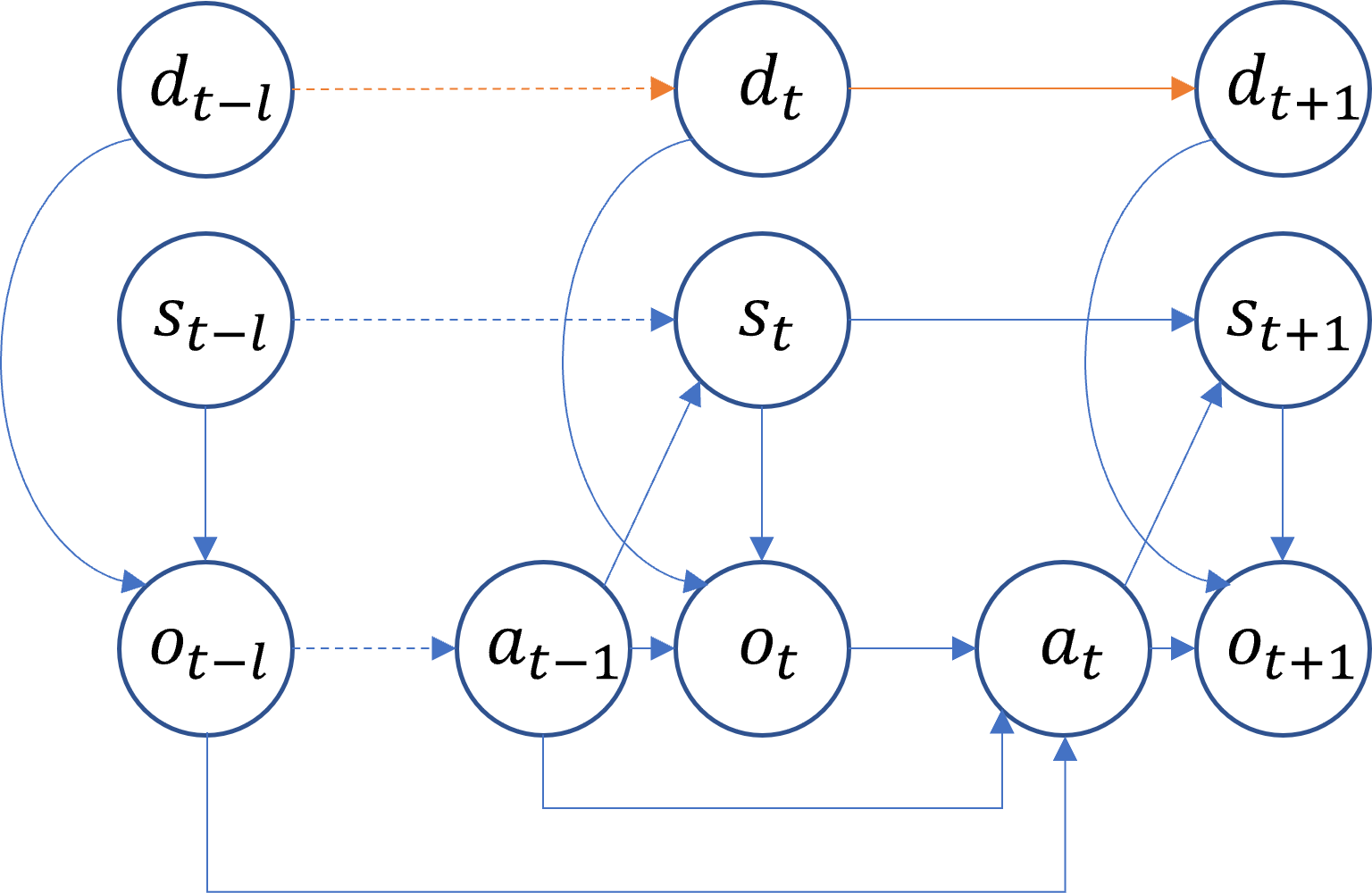}%
\label{fig:calsual_diagram}}
\hfil
\subfloat[]{\includegraphics[width=0.45\textwidth]{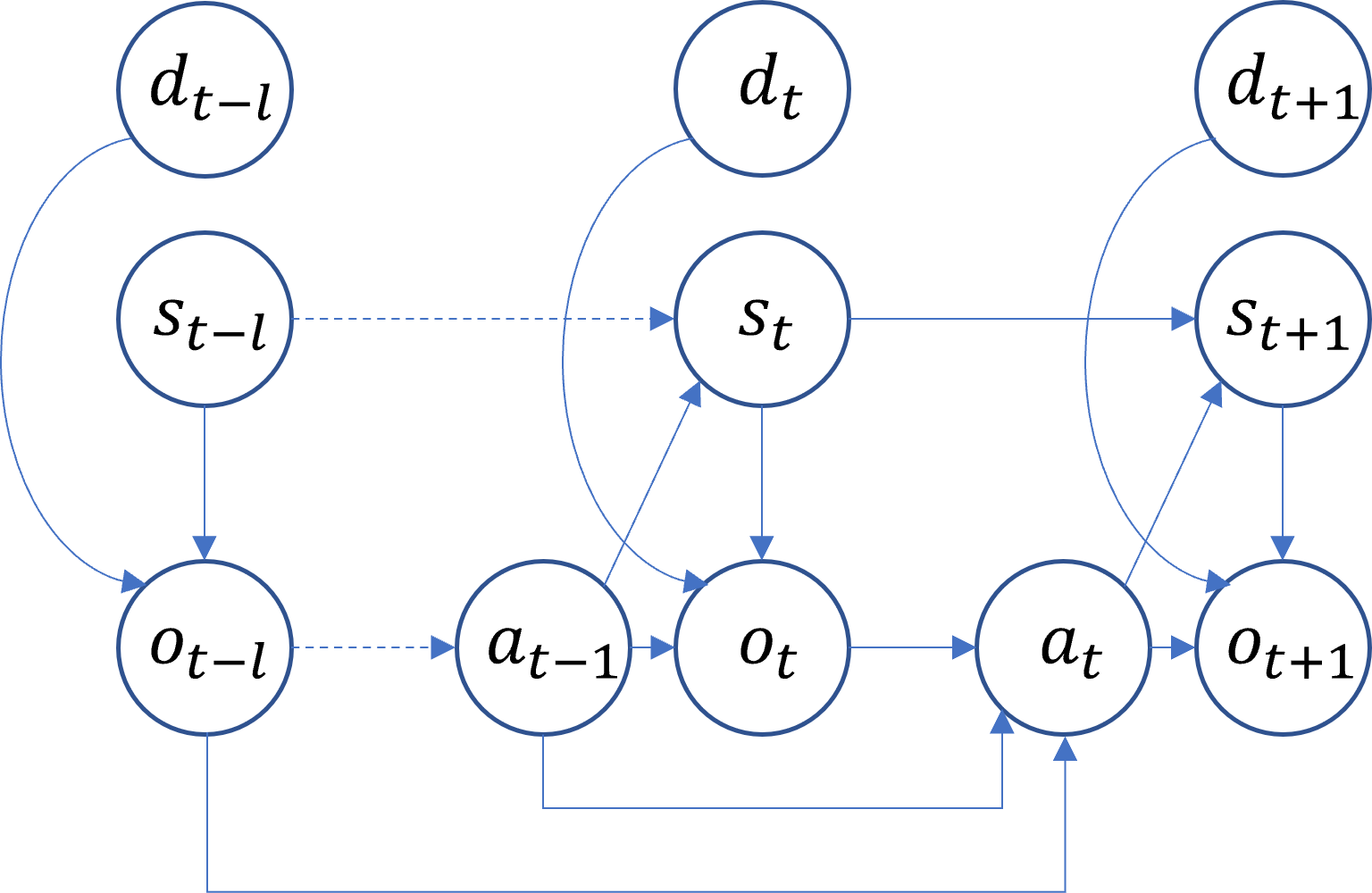}%
\label{fig:calsual_diagram2}}
\caption{Causal diagram. (a) With dynamic disturbance. (b) With non-dynamic disturbance}
\label{fig:causal}
\end{figure*}

When the decision-making is performed based on information on the trajectory, the causal system is illustrated in Fig.\ref{fig:causal}. In this work, the considered disturbances can be put into two categories. The first case is where the disturbance can be described by a dynamic model (Fig. \ref{fig:calsual_diagram}). In this case, the trained agent may be able to learn a model comprising the system and the disturbance. The second case is where the disturbance does not have a temporal correlation (Fig. \ref{fig:calsual_diagram2}). In this case, the agent should learn to eliminate the effect of the disturbance to reveal the true state.
The agent trained in the environment, as categorized in Fig. ~\ref{fig:calsual_diagram}, might demonstrate generalizability in an environment where the existing disturbance has a temporal correlation, but not in the other one. This will be demonstrated in Section \ref{discussion}.

\subsection{POMDP Formulation}
In general, a POMDP model is defined as a 6-tuple $\langle \mathcal{S, A, O, T, Z, R} \rangle$ \cite{kurniawati2021partially}, where 
\begin{itemize}
    \item $\mathcal{S}$ is the state space,
    \item $\mathcal{A}$ is the action space,
    \item $\mathcal{O}$ is the observation space,
    \item $\mathcal{T}(s_t,a_t,s_{t+1})$ is the state transition which is a conditional probability distribution $P(s_{t+1}\vert s_{t}, a_{t})$,
    \item $\mathcal{Z}(s_{t+1}, a_t, o_{t+1})$ is the observation function which is a conditional probability distribution $P(o_{t+1}\vert s_{t+1}, a_{t})$,
    \item $\mathcal{R}$ is an immediate reward function which represents $\mathcal{R}=\mathbb{E}\left[r_{t+1}\vert s_t, a_t \right]$.
\end{itemize}
When observations are corrupted by additive dynamic disturbances $d_t$, the observation function $\mathcal{Z}$ is not static. Namely, the observation function can be described as $P(o_{t+1}\vert s_{t+1}, a_{t}, d_{t})$. In addition, if the disturbance is modeled and it is independent of the states and action, a new disturbance transition function, which represents a conditional probability distribution could be introduced.

\subsection{Information States, Belief States in POMDP}
In the context of dynamic programming, the optimal controller in POMDP can be separated into two parts; $(a)$ an estimator and $(b)$ an actuator \cite{bertsekas2012dynamic,kurniawati2021partially}. 
The information available at time $t$ is called complete information state $I_t^C$ \cite{hauskrecht1997planning} which consists of all historical observation and action i.e., $I^C_t=(o_{0:t}, a_{0:t-1})$, where $o_{0:t} = \left
\{o_0, \dots , o_t\right\}$ and $a_{0:t-1} =  \left
\{a_0 ,\dots, a_{t-1}\right\}$. In the case of POMDP, the estimator provides the conditional expectation $\hat s_t  = \mathbb{E}(s_t\vert I^C_t)$. The challenge of utilizing the complete information states is that the size of the data expands as time elapses. To resolve this issue, quantities known as sufficient statistics are normally considered. The sufficient statistic has a smaller dimension and preserves the essential content of $I^C_t$. The sufficient statistic is generated at the estimator and the actuator generates control inputs to the system based on the sufficient statistic. In the standard POMDP model, a belief state $b_t$ is commonly used as a sufficient statistic. The belief state assigns conditional probabilities of every possible state \cite{hauskrecht2000value}.

As illustrated in \cite{kurniawati2021partially}, the policy of the agent receives the belief states to make optimal decisions.
The belief states could be recursively updated via Bayesian inference. 
The update of belief states can be described as (\ref{equ:beliefupdate}).
\begin{flalign}\label{equ:beliefupdate}
    b_{t+1}(s_{t+1}) &= P(s_{t+1}\vert o_t, a_t, b_t)  \\ \nonumber
    &= \frac{P(o_t\vert s_{t+1}, a_t, b_t)P(s_{t+1}\vert a_t, b_t)}{P(o_t \vert a_t, b_t)} \\ \nonumber
     & = \frac{\mathcal{Z}(s_{t+1}, a_t, o_t)}
     {\sum_{s_{t+1}\in S}\mathcal{Z}(s_{t+1}, a_t, o_t)}\\ \nonumber
     & \qquad\qquad\qquad \frac{\sum_{s_t\in \mathcal{S}}\mathcal{T}(s_t,a_t,s_{t+1})b_t(s_t)}
     {\sum_{{s_t}\in S} \mathcal{T}(s_t,a_t,s_{t+1})b_t(s_t)}
\end{flalign}

\subsection{Identification of Transition Model}
The goal of RL is to maximize the future cumulative reward $G_t$ by learning an optimal policy $\pi$. In model-based RL, this problem is tackled by decomposing it into two sub-problems: learning and planning. In the learning phase, the transition model is explicitly derived from experience, while during planning, the value function is established based on this learned model. On the other hand, in model-free RL, the agent is directly trained from experience, thus the transition model and its update operation formulated in (\ref{equ:beliefupdate}) are not required. Even though the transition model is not explicitly used in model-free RL, the concept is still applicable as described in Section \ref{background}. In this case, the model is internally generated based directly on the collected data. Since actions are part of the transition model and thus part of the recursive update, including action sequences in model-free RL can enhance the robustness of optimality in POMDP situations.

Regardless of whether the complete information state $I^C_t$ or belief state $b_t$ is used, the identification of the dynamic transition model depends not only on the historical information of the observation but also that of the action. This is because both have causal connections with the current observation, as illustrated in Fig.~\ref{fig:calsual_diagram}. As we are considering the environment where the observation is disturbed, our discussion is focused on generating internal representations of $s_t$ which is symbolized by $s^{*}_t$ in the rest of this paper. The key idea is to formulate a structure of an RL agent that can dynamically adapt the state representation $s^{*}_t$ on given disturbances.

\subsection{LSTM to Reflect the Sequence in the Internal Representation $s^{*}_t$}
When the model of the system is not available, $s^{*}_t$ could be generated via RNN by processing the sequential input. $s^{*}_t$ contains the identified information regarding the pattern of the dynamic model of the system and the disturbance or distilled information which is less affected by the disturbance. Most of the existing RL algorithms with RNNs only utilized the sequential information of observation and there are only a few cases taking action sequences as input. We suggest utilizing the sequence of action based on the above discussion in this study. Therefore, the effect of sequential action inclusion with the LSTM-TD3 algorithm \cite{meng2021memory} is investigated in the next section. 

\section{Action Inclusion}\label{actionInclusion}

In the previous section, we have discussed that the belief state evolves based on the observation and action on the trajectory. In this section, the impact of the inclusion of the action sequence is demonstrated using LSTM-TD3. Also, the effect of the length of the sequence for the individual disturbance condition is investigated. 

\subsection{Length of Sequence}
When the partial trajectory is taken as the input sequence, the selection of its length $l$ can be a design parameter as in LSTM-TD3. In general, RL algorithms with an RNN layer require repeating either entire or partial episodes during the update in order to train the network via Back Propagation Through Time (BPTT). When entire episodes are repeated in training, it can be considered that the agent is trained using complete information $I^C$. As stated in Section \ref{keyidea}, the size of $I^C$ increases as time elapses. Thus, repeating entire trajectories to update the network is a time-consuming process. In case of the on-policy algorithm with RNN, utilizing $I^C$ might be a more natural approach since it follows the same trajectory during the training. Several works have explored the methods combining RNN with on-policy algorithms (e.g., PPO \cite{gaudet2019adaptive} and A3C \cite{mnih2016asynchronous}). However, on-policy algorithms are much less sample-efficient compared to off-policy algorithms. 

For state value function based algorithms, parameter sharing is a common technique where the actor and critic share the same network parameters and only the last layer of each network is different. However, this technique is not as common for action value function based algorithms like TD3 since the parameter sets are different between actor and critic networks. When parameter sharing is not used, the same trajectory needs to be repeated multiple times to update multiple networks.

If the whole episode is repeated, the benefit provided by random sampling can be lost in the experience replay which is the mechanism for improving data efficiency and stability. Experience replay effectively minimizes the correlations between samples \cite{lillicrap2015continuous} so that the data distribution becomes closer to independent and identically distributed (i.i.d.) condition which is preferable in RL. In \cite{heess2015memory}, it is concerned that if the whole trajectory is sampled together, the bias in the learned policy might become more severe in the update because the state distribution under the current policy no longer corresponds to the one from the replay buffer.

If a few steps of sequential data are randomly sampled as a unit, the sample correlation issue could be mitigated. This approach captures a shorter history of fixed length in the environment. LSTM-TD3 offers the opportunity to define the length of the sequence. The agent should be provided with enough length of sequential data so that it can capture the characteristics of the system and the disturbance. In the experiment in this section, the relationship between several characteristics of the disturbance and the length of sequence provided to the network is investigated.

\subsection{Experiment}\label{experiment}
To validate the discussion in Section \ref{keyidea} and investigate the effect of including action sequence, we compare learning trajectories of the LSTM-TD3 algorithm between the cases with or without the past action sequence in the first input channel (see Fig. \ref{fig:lstmtd3}) in classic control “Pendulum” environment from OpenAI gym. In this environment, the observation consists of three elements $\left[x,y,\dot\theta \right]$, where $x$ and $y$ represent the position of the free end of the pendulum and $\dot\theta$ is its angular velocity. The action is defined as the torque to be added and the agent learns to keep the pendulum upright from a random initial condition. The definition of the “Pendulum-v0” environment is summarized in Table \ref{tab:Pendulum-v0}.\\
\begin{table}[H]
\begin{center}
\caption{“Pendulum-v0” environment definition} \label{tab:Pendulum-v0}
\begin{tabular}{ |c|c|c|c|  }
\hline 
 action & torque & min $-2.0$ & max $2.0$ \\
\hline
\multirow{3}{*}{observation} & $x=\cos{\theta}$ & min $-1.0$ & max $1.0$ \\
& $y=\sin{\theta}$ & min $-1.0$ & max $1.0$ \\
& angular velocity & min $-8.0$ & max $8.0$\\
\hline
reward  &\multicolumn{3}{|c|}{$r = -(\theta^2 + 0.1\Dot{\theta}^2 + 0.001 (torque)^2)$} \\
\hline
termination  &\multicolumn{3}{|c|}{at $200$ time steps} \\
\hline    
\end{tabular}
\end{center}
\end{table} 
The motivation for this work is to develop an algorithm that robustly provides optimality in the presence of disturbance. As summarized in Table \ref{tab:scenario}, we classified disturbance under condition into five types: “temporal bias”, “temporal sinusoidal wave”, “random sinusoidal wave”, “noise” and “hidden” conditions. In the “temporal bias” and the “temporal sinusoidal wave” cases, the observation of $x$ and $y$ are disturbed by additive constant biases and sinusoidal waves for specific lengths and times at random time steps. In the “random sinusoidal wave” case, the sinusoidal wave is randomly configured at each episode and all three elements are disturbed. In the “noise” case, zero-mean Gaussian noise with standard deviation $\sigma$ is added to each element of the observation during the entire episode. In the “hidden” condition, $\dot\theta$ is removed from the observation during the whole episode reducing the dimension of the observation to two. Scenarios 1, 2, and 3 in Table \ref{tab:scenario} are the cases where the disturbance has its dynamics and its dynamic patterns could be identified from the observation. The other two remaining cases have no dynamics in the disturbance. The networks should rather work on recovering the dynamics of the system by eliminating the effect of the disturbance from the observation than establishing a model that includes both the system and disturbance.
\begin{table*}[ht]
\centering
\caption{Test scenario}
\label{tab:scenario}

\begin{tabularx}{\textwidth}{p{1.5cm} p{2.5cm} X}
\hline\hline
Scenario & Name & Description \\
\hline

1 & temporal bias &
Disturbance defined by $A\in[0.5,1.0]$ appearing randomly is added to observation in $x$ and $y$, lasting $3$ time steps for $10$ times per episode. \\

\hline
2 & temporal sinusoidal wave &
Disturbance defined by $A\sin(2\pi t/T)$ ($A=1$, $T=70$) starting at randomly chosen time steps for period $T$, added to observation $x$ and $y$. \\

\hline
3 & random sinusoidal wave &
Disturbance randomly defined in every episode by $A\sin(2\pi t/T)$ ($0.5\leq A \leq 2$, $10\leq T\leq 100$) starting at a random time step for period $T$, added to all three observation elements. \\

\hline
4 & noise &
Zero-mean Gaussian noise with standard deviation $\sigma \in [0.5,1.0]$ constantly added to all three observation elements. \\

\hline
5 & hidden &
Angular velocity $\dot{\theta}$ is removed from observation. \\

\hline\hline
\end{tabularx}
\end{table*}

\subsection{Results and Analysis}
\begin{figure*}[ht]
    \centering
    \includegraphics[width = \textwidth]{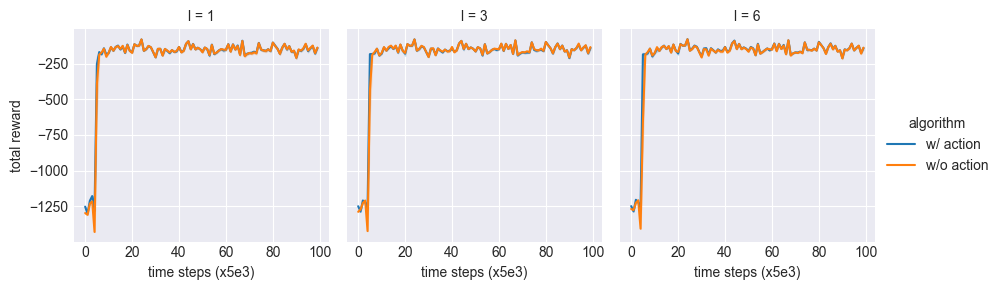}
    \caption{Results for Fully Observable MDP Case}
    \label{fig:action_MDP}
\end{figure*}
\begin{figure*}[ht]
    \centering
    \includegraphics[width = \textwidth]{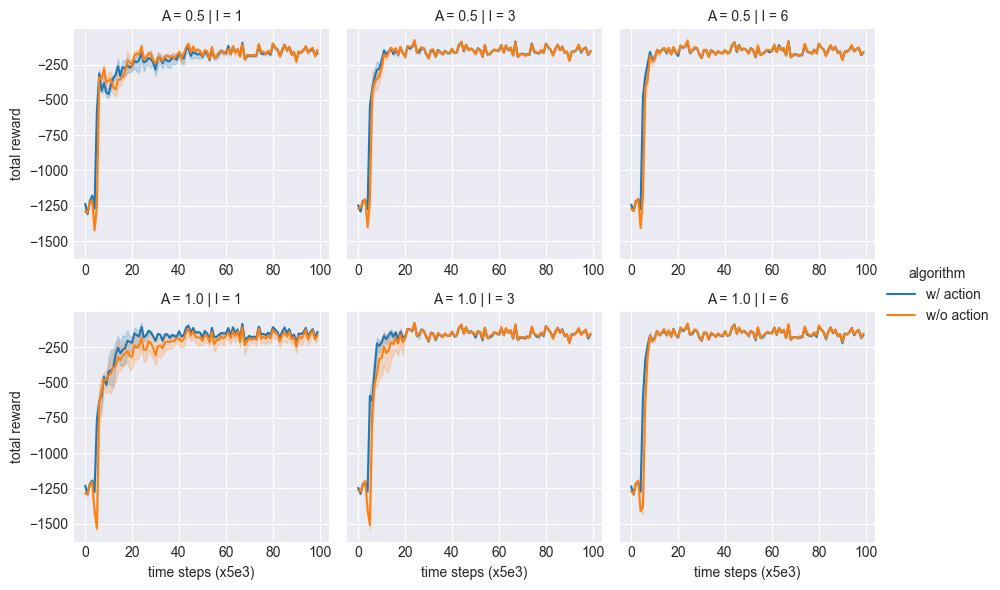}
    \caption{Results for Scenario 1 - Temporal Bias}
    \label{fig:action_bias}
\end{figure*}
\begin{figure*}[ht]
    \centering
    \includegraphics[width = \textwidth]{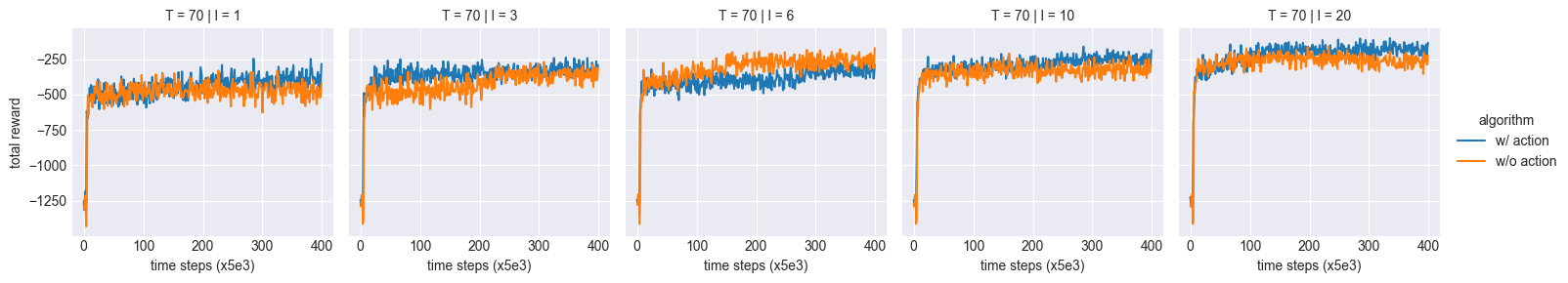}
    \caption{Results for Scenario 2 - Temporal Sinusoidal Wave}
    \label{fig:action_sin}
\end{figure*}
\begin{figure*}[ht]
    \centering
    \includegraphics[width = \textwidth]{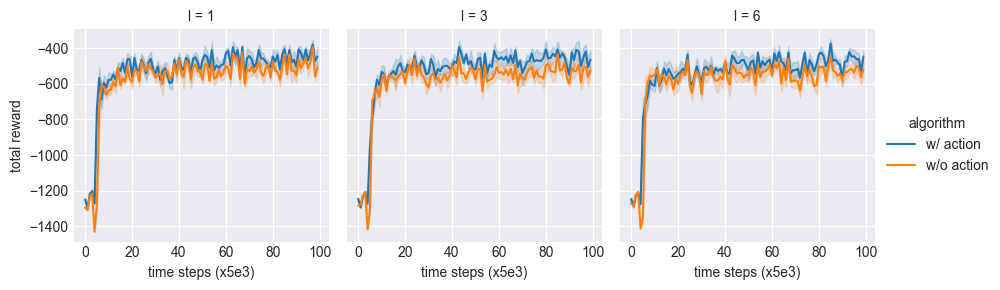}
    \caption{Results for Scenario 3 - Random Sinusoidal Wave}
    \label{fig:action_sin4}
\end{figure*}
\begin{figure*}[ht]
    \centering
    \includegraphics[width = \textwidth]{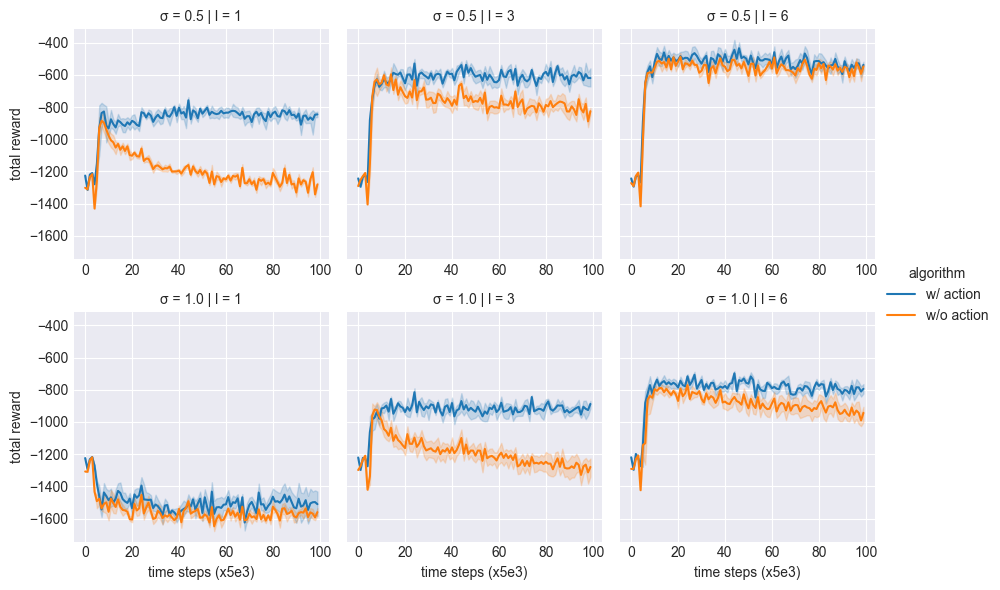}
    \caption{Results for Scenario 4 - Noise}
    \label{fig:action_noise}
\end{figure*}
\begin{figure*}[ht]
    \centering
    \includegraphics[width = \textwidth]{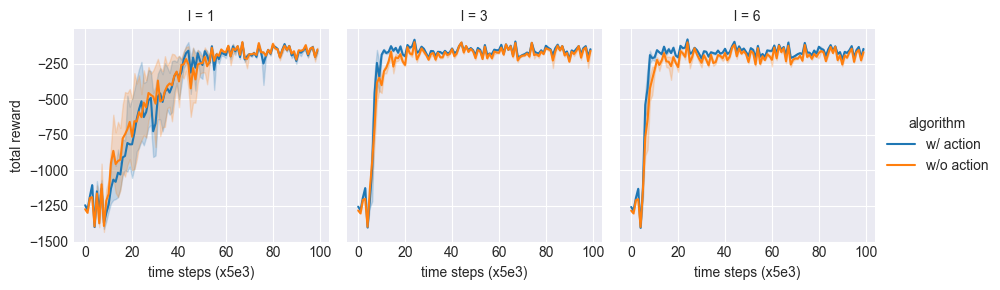}
    \caption{Results for Scenario 5 - Hidden}
    \label{fig:action_hidden}
\end{figure*}
Fig. \ref{fig:action_MDP} to Fig. \ref{fig:action_hidden} show normalized learning trajectories averaged over 5 trials of the LSTM-TD3 algorithm in the discussed scenarios. The orange lines, labeled “w/o action”, show the case in which sequential action information is not considered. Specifically, the first input channel of actor and critic networks processes only sequential $l-1$ length of historical observation $o_{t-l:t-1}(=o_{t-l}, \dots,o_{t-1})$. The blue lines, labeled “w/ action”, show the case where the historical action $a_{t-l:t-1}$ is used as additional information. The columns represent the length of the history to be processed. Fig. \ref{fig:action_MDP} and Fig. \ref{fig:action_bias} illustrate the results of the MDP case and the “temporal bias” case respectively. In the MDP case, LSTM-TD3 with or without action performs the same. Also, the final performance at the end of training in the “temporal bias” case did not show obvious differences. In this condition, a constant value is added for three time steps that randomly start appearing 10 times. Since the observation is biased by a constant value, the observation is still almost unique. Therefore, it can be considered to be similar to the MDP condition. 

Fig. \ref{fig:action_sin} shows the results of the “temporal sinusoidal wave” scenario. As $l$ gets longer, the robustness improves. The wavelength of the defined sinusoidal wave is $70$. Longer $l$ helps to identify the system dynamics and the disturbance dynamics. Also, the case included the action sequence as the input to the agent achieved higher total rewards. The result indicates the impact of information selection on robustness. 

Fig. \ref{fig:action_sin4} represents the learning trajectory of the “random sinusoidal wave” scenario. Establishing the observation function in this scenario is much harder than in the “temporal sinusoidal wave” due to the infinite variety of sinusoidal waves. It requires the agent to dynamically adapt to the environment based on the causality of action and observation and hence, the network with action sequence achieves greater optimality. 

Fig. \ref{fig:action_noise} illustrates the learning trajectory under the “noise” condition. This scenario is more challenging than the “temporal sinusoidal wave” since there are no specific dynamics in the disturbance and all of the elements of the observations are constantly disturbed. Although the network considering sequential action information performs better, the performance degrades with the larger variance of the noise. The performance improves with longer $l$. Longer sequential data contributes to extracting the system state transition characteristic and eliminating the effect of the noise in LSTM. 
In the white noise case, some data randomly contain less noise. By focusing on these data and putting less focus on more noisy data, the underlying pattern can be restored. When the length of the input sequence is longer, it has more chance to have informative data that is less affected by noise. In addition, as more episodes are experienced, degradation of achieved total reward was observed in the “noise” scenario. This would be due to the occurrence of overfitting. It might be prevented by changing the network configuration, e.g., reducing the number of parameters, and adding dropout layers.

Fig. \ref{fig:action_hidden} presents the result of the “hidden” scenario. The final performance is the same regardless of the action inclusion and length of the sequence. In this environment, only current observation $o_t$ and one step before $o_{t-1}$ should be sufficient to restore $\dot \theta$ since $\theta$ is updated as $\theta(k+1) \leftarrow \theta (k) + \dot\theta (k+1)dt$. The formula only requires the previous observation to restore $\dot{\theta}$ from the equation and therefore the performance is not affected by $l$. However, the past action information is making the learning process more efficient. The network does not have this mentioned interpretation and the reward signal is mixed with $\theta$, $\dot \theta$, and torque. The inclusion of action information could make learning efficient because the network learns to process the dynamics under more specific conditions. The most reasonable network structure to capture this interpretation might instead just consist of fully connected layers by having the input of $\left[x_t, y_t, x_{t-1}, y_{t-1} \right]$ because the sequential process is not required.

\section{Network Architecture for Sequential Data Handling}\label{structure}
In Section \ref{actionInclusion}, we demonstrated that including the action sequence can improve the robustness of learning in dynamic environments.
In this section, let us discuss the structure based on interpreting the reconstructed state representation via LSTM. 
\subsection{Interpretation of LSTM-TD3}
The LSTM-TD3 has two input channels in both actor and critic networks. 
They could be interpreted as the first input channel to generate $s^*_{t-1}$ from complete information $I^C_{t-l:t}$ taken from the memory. Then, the $s^*_{t-1}$ is updated to $s^*_{t}$ by combining the current observation information from the second channel at the actor. Our hypothesis is that it might not be necessary to have a double-headed structure as presented in Fig. \ref{fig:lstmtd3} because the information up to the current time step can be treated as one sequence according to the generation of the information states.

\subsection{Modified LSTM-TD3}
\begin{figure*}[!t]
\centering
\subfloat[]{\includegraphics[width=0.45\textwidth]{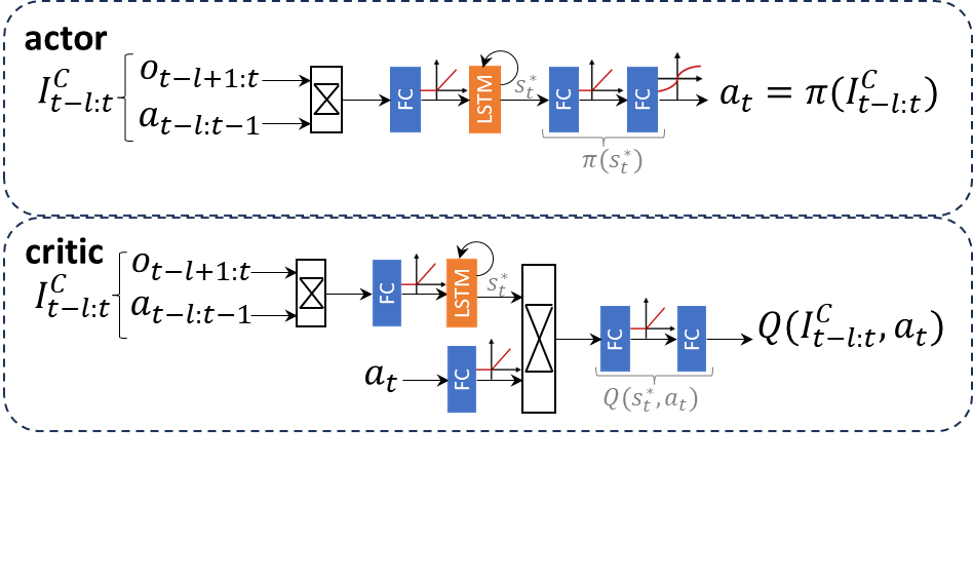}%
\label{fig:1ha2hc}}
\hfil
\subfloat[]{\includegraphics[width=0.45\textwidth]{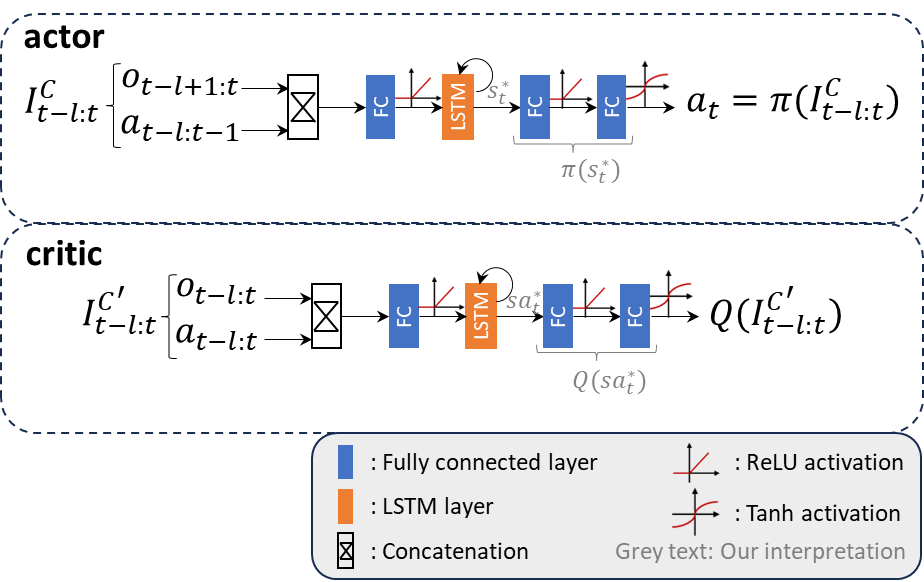}%
\label{fig:1ha1hc}}
\caption{Modified LSTM-TD3 structure. (a) LSTM-TD3\textsubscript{1ha2hc}. (b) LSTM-TD3\textsubscript{1ha1hc}.}
\label{fig:structure}
\end{figure*}

Based on this hypothesis, this paper proposes an actor network with a single input channel as illustrated in Fig. \ref{fig:structure}. This input channel processes $I^C_{t-l:t} =(o_{t-l}, \dots , o_t, a_{t-l-1} ,\dots, a_{t-1})$. Internal state representation $s^*_t$ is generated in the process of the LSTM layer and action is decided based on $s^*_t$ by the succeeding layers. For the critic network, two types of structure are considered. LSTM-TD3\textsubscript{1ha2hc} (1 headed actor and 2 headed critic) in Fig. \ref{fig:1ha2hc} has a clearer interpretation for $s^*_t$ to the information states. $s^*_t$ is generated in the same way as actor network at the first input channel and it is concatenated with the processed $a_t$ at another input channel. The last two layers can be understood as $Q(s^*_t, a'_t)$. On the other hand, LSTM-TD3\textsubscript{1ha1hc} has single input channel and modified complete information  $I^{C'}_{t-l:t} =(o_{t-1}, \dots , o_t, a_{t-l} ,\dots, a_{t})$ is taken as input. $I^{C'}_{t-l:t}$ differs from the generation of $s^*_t$ based on the Bayesian inference and hence, the output from the LSTM layer is described as $sa_t$ in the figure. However, the definition of action value function $Q$ is the expected return at $s_t$, taking action $a_t$ when following policy $\pi$. When the critic network is given with the trajectory, $I^{C'}_{t-l:t}$, which is the combination of $o_t$ and $a_t$ from $\pi$, it should be able to still generate the expected return for $\pi$.  

Unlike the original LSTM-TD3, LSTM-TD3\textsubscript{1ha2hc} and LSTM-TD3\textsubscript{1ha1hc} do not have an explicit intention to prioritize the current information. These proposed algorithms follow the update of the belief state and allow LSTM to evaluate the importance of the data. Moreover, since the system model does not change in any time step, all data in a sequence should be treated in the same path. Training the agent with separate sequences might be more complicated because the data is processed differently and combined which might be unnecessary according to the belief state update. Therefore, LSTM-TD3\textsubscript{1ha2hc} and LSTM-TD3\textsubscript{1ha1hc} are expected to show better robustness.

\subsection{H-TD3 Algorithm}
The previously discussed algorithms generate $s^*_t$ in actor and critic networks independently by taking complete information at each network. Particularly, we have introduced the interpretation where $s^*_t$ is produced in the LSTM layer at actor network. The research question to be also investigated is whether or not it is necessary to have a different representation of $s^*_t$ between the two networks. If $s^*_t$ generated from the actor network can be shared with the critic network, it will drastically reduce the computational costs. Because TD3 has double critics, $I^C_{t-l:t}$ is processed four times at each critic network during the update. Hence we propose a new algorithm, named H-TD3 (hidden-state-based TD3) in which the critic networks are trained based on the shared $s^*$ from the behavior actor network during the exploration. Typically, the RL algorithms with off-policy actor-critic style do not share the network, unlike on-policy algorithms. In H-TD3, the state of the LSTM of the actor network is shared with the critic network to initialize LSTM state to avoid processing the data as sequence under the assumption that the LSTM state of the actor can contain the summarized sequential information.

\begin{figure}[h]
\begin{center}
\noindent
  \includegraphics[width = \linewidth]{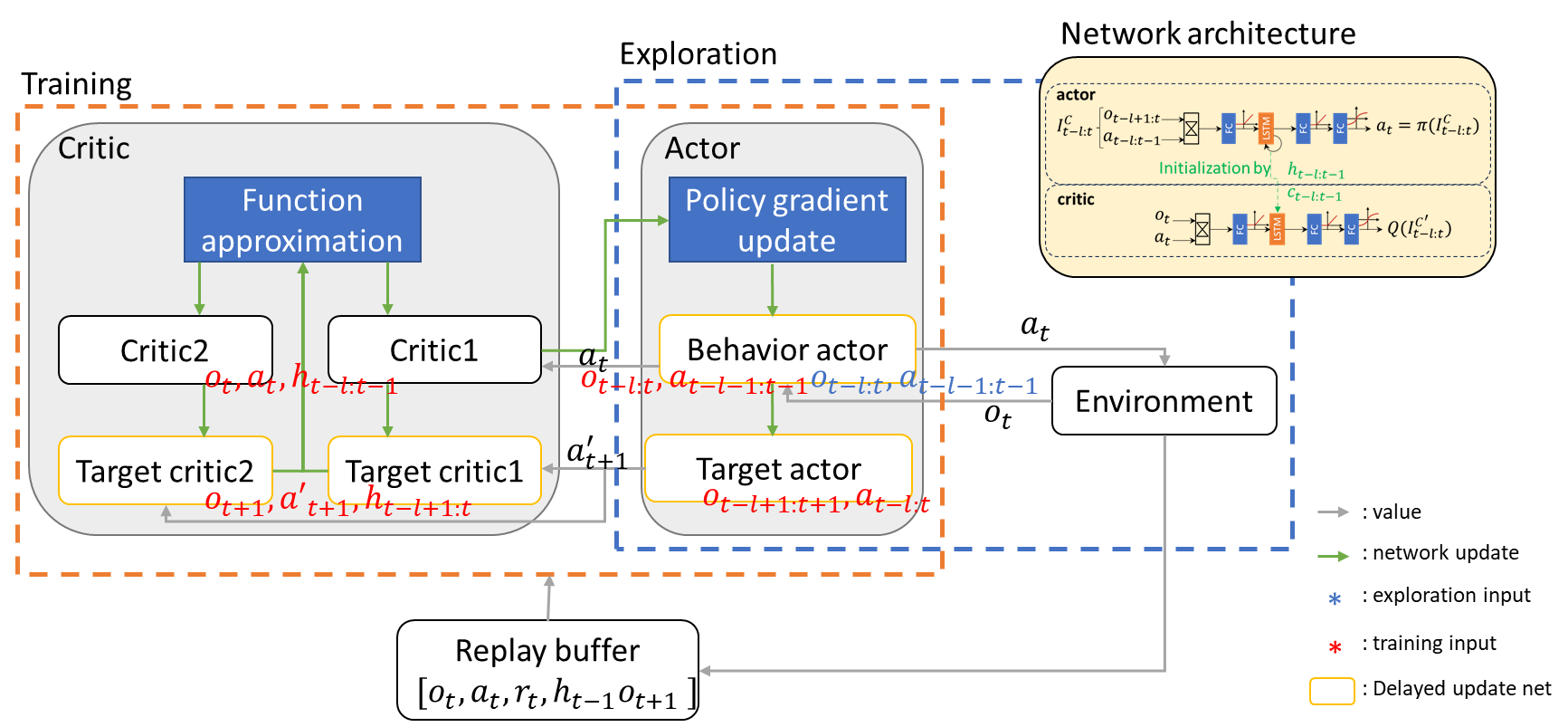}
  \caption{H-TD3 schematics}\label{fig:htd3structure}
\end{center}
\end{figure}

Here, we consider the same network structure with LSTM-TD3\textsubscript{1ha1hc} which has an LSTM layer with a single input channel for the critic network as it is illustrated on the right side of Fig. \ref{fig:htd3structure}. The main diagram of Fig. \ref{fig:htd3structure} describes the data flow of the H-TD3 algorithm. During training, the behavior actor and environment interact and the collected data is stored in the replay buffer. In addition to the normal data set, the hidden state $h_t$ and the cell state $c_t$ of the LSTM cell are stored. These states are generated at the behavior actor network when LSTM processed the sequence of $a_{t-l-1:t-2}$ and $o_{t-l:t-1}$. The critic network is trained based on this tuple. Critic networks take $a_t$ and $o_t$ as input data as the original TD3 algorithm. However, the LSTM cell is initialized with the stored $(h_{t-l:t-1}, c_{t-l:t-1})$. In this way, input data $a_t$ and $o_t$ are treated as if they formed a continuous sequence with $a_{t-l-1:t-2}$ and $o_{t-l:t-1}$, without repeating the entire trajectory between $t-l$ and $t$. In the same way, at the target critics, the target Q-values are calculated based on input data $a'_{t+1}$ and $o_{t+1}$ with initialized LSTM state with $(h_{t-l+1:t}, c_{t-l+1:t})$. The critic network is required to learn parameters based on $(a_t, o_t, h_{t-l:t-1}, c_{t-l:t-1})$. The criticism of this approach may be the dismissed action $a_{t-1}$ and $a_{t}$ at critic and target critic networks respectively. However, if the hidden states $(h_{t-l:t}, c_{t-l:t})$ of LSTM after processing the entire sequence $a_{t-l-1:t-1}$ and $o_{t-l:t}$ are used to initialize critic networks, $o_{t}$ is counted twice and it would break the sequence at the critic. Another possibility could be to limit the input just to $a_{t}$, but changing the character of input data to LSTM for the given $(h_{t-l:t}, c_{t-l:t})$ in critics would add difficulty in learning. During the development, we observed the degradation of performance in both cases compared to using  $(h_{t-l:t-1}, c_{t-l:t-1})$.

Considering the developed LSTM-TD3\textsubscript{1ha2hc} and LSTM-TD3\textsubscript{1ha1hc}, the output data from LSTM in actor network is the last hidden states $h_t$ after processing $l$ length of the sequential input. After the LSTM layer, $h_t$ is carried to the following last two layers. Hence, the simplest method for the critic networks to learn based on the shared $s^*$ would be taking $h_t$ as input data at the critic network in addition to the current observation. However, this operation would extend the input dimension excessively as the dimension of the hidden state is usually much larger than that of the input. It increases the difficulty to train the network and the risk of failure of learning as known as the “curse of dimensionality”. Training network with larger input dimensions increases the number of parameters drastically, leading to the much greater computational complexity of training. Also, ending up with over-fitting is a common issue. Therefore, it is more scalable to share the hidden and cell states produced in the actor network to initialize the hidden state of the critic network as in the proposed H-TD3 algorithm.

\section{Experiment}\label{discussion}
\subsection{Simulation Set-Up}
In this section, the algorithms developed in Section \ref{structure} are tested in the “Pendulum” environment with disturbance conditions discussed in Section \ref{actionInclusion}. The applied disturbance conditions are as follows:
\begin{itemize}
    \item “temporal sinusoidal wave” with $\sin\frac{2\pi}{70}t$
    \item “random sinusoidal wave”
    \item Zero-mean Gaussian “noise” with $\sigma=0.5$
    \item “hidden”
\end{itemize}
\subsection{Results}
\begin{figure*}[ht]
    \centering
    \includegraphics[width = \textwidth]{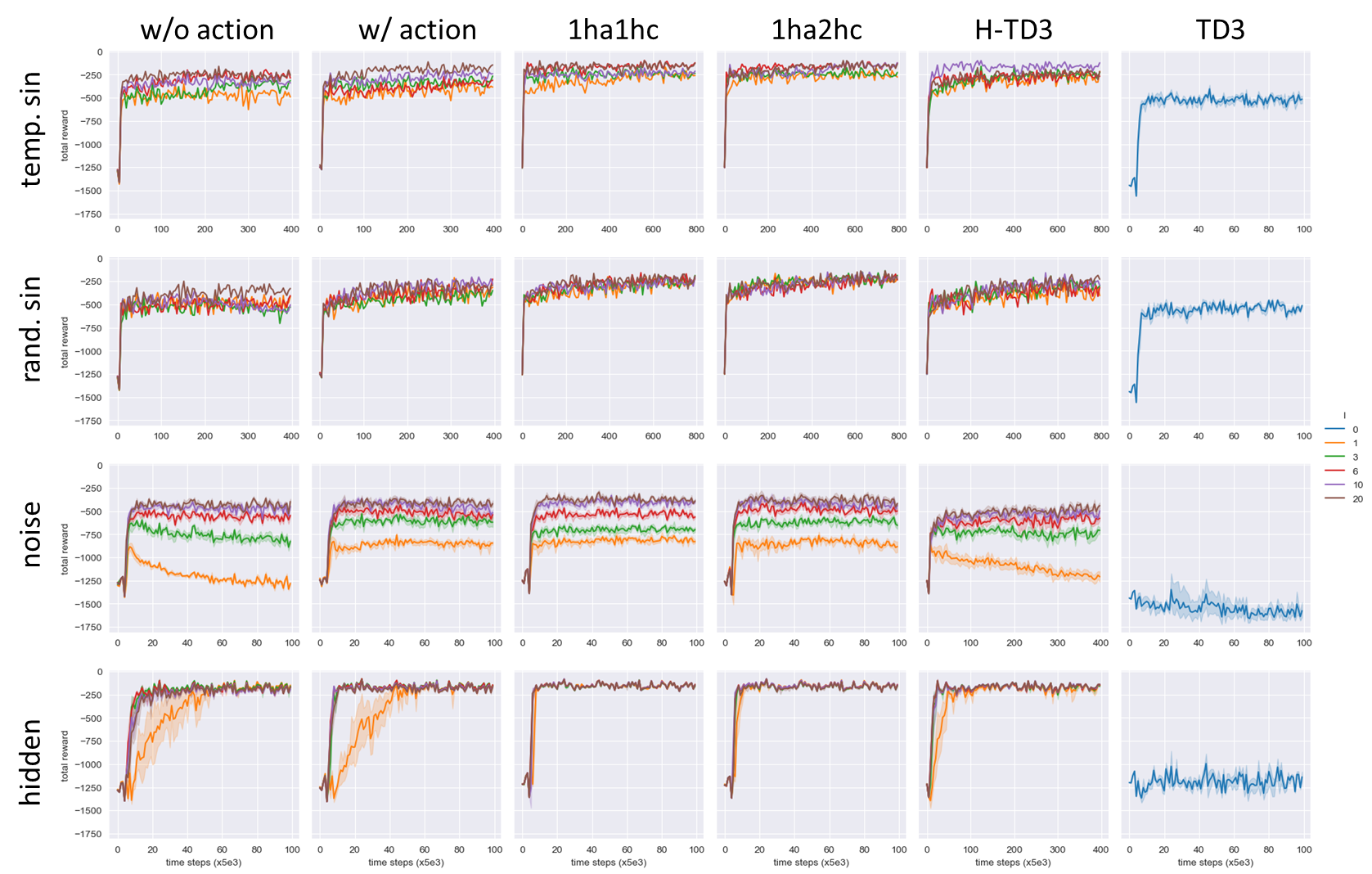}
    \caption{Learning trajectory}
    \label{fig:all_traj}
\end{figure*}

In Fig. \ref{fig:all_traj}, the learning trajectories of the proposed algorithms are depicted, with trajectories being averaged over 5 trials. We experiment with various time windows $l$, specifically $1, 3, 6, 10, 20$. In the array of figures, each row corresponds to a specific disturbance condition, while each column represents a different algorithm. The algorithms compared are original LSTM-TD3 without past action sequence, LSTM-TD3 with past action sequence, LSTM-TD3$_{1ha1hc}$, LSTM-TD3$_{1ha2hc}$, H-TD3 and original TD3 from the left column. The disturbances are “temporal sinusoidal wave”, “random sinusoidal wave”, “noise” and “hidden” from the top. The number of iterations varies depending on the disturbance conditions. 

\begin{figure*}[ht]
    \centering
    \includegraphics[width = \textwidth]{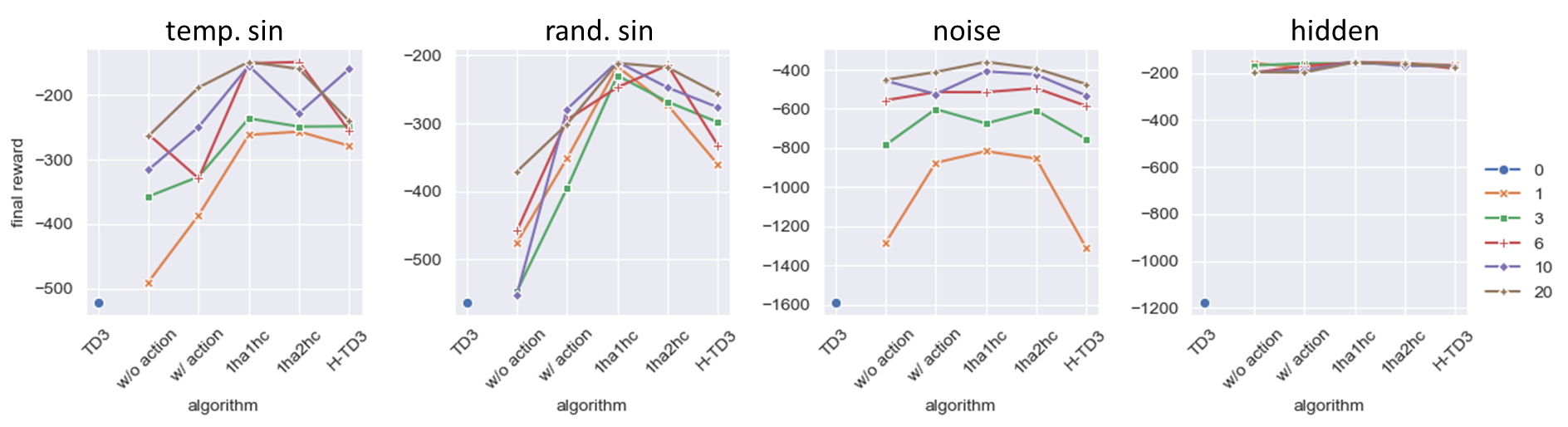}
    \caption{Final total reward}
    \label{fig:endperformance}
\end{figure*}

Fig. \ref{fig:endperformance} shows the normalized total reward of the last ten records by taking algorithm types on the x-axis. Each figure describes the result of “temporal sinusoidal wave”, “random sinusoidal wave”, “noise” and “hidden” conditions respectively. The lines are drawn to compare the effect of the time window $l$.

\paragraph{Overall Analysis}
Compared to TD3, all algorithms using an LSTM layer demonstrated superior performance in the tested POMDP conditions. Particularly, in the “noise” and the “hidden” cases, TD3 struggled to learn. The “noise” environment generates the reward based on the noisy observation. Without access to a sequence, it is difficult to learn since the observation values are not unique to the reward signal. Also, in the “hidden” scenario, a model without an LSTM layer struggled to recover information related to angular velocity which is part of the reward function.

As observed in Section \ref{actionInclusion}, all algorithms incorporating action sequences performed better than LSTM-TD3 without an action sequence. The variants LSTM-TD3 (LSTM-TD3$_{1ha1hc}$, LSTM-TD3$_{1ha2hc}$) generally performed better than the original LSTM-TD3 with action sequence. From the results, it can be understood that it is better to treat the past sequence and current data as one sequence, unlike LSTM-TD3. LSTM-TD3$_{1ha1hc}$ exhibited the best robustness in the optimality of all algorithms. The simplicity of the model architecture could have helped the network to learn. H-TD3 algorithm delivered a comparable result with LSTM-TD3 with action case except for the “noise” case. It confirms that it is possible to learn from shared hidden states as demonstrated by the H-TD3 results. However, H-TD3 exhibited slower learning trajectories for a given number of iterations than the other algorithms in Fig. \ref{fig:all_traj} due to the critic's dependence on the actor network to generate the hidden states and cell states. Interestingly, the results of the “noise” case in H-TD3 were significantly degraded and mostly worse than LSTM-TD3 without action sequence. In the H-TD3 algorithm, $a_{t-1}$ is not included in the information. The effect of omitting this information could become prominent in some conditions as in the “noise” case.

\begin{figure}[h]
\begin{center}
\noindent
  \includegraphics[width = 0.8\linewidth]{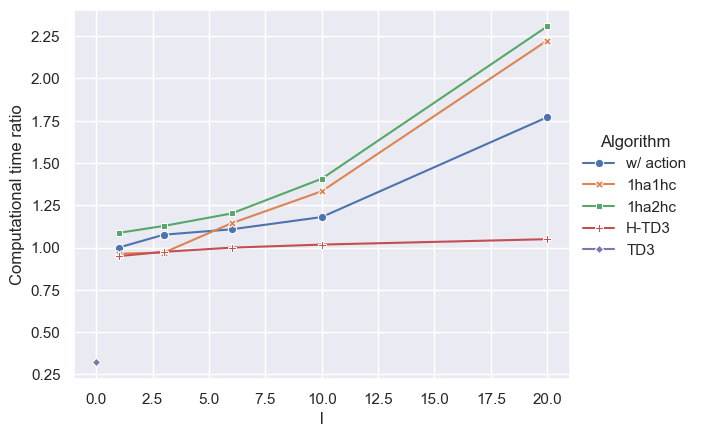}
  \caption{Iteration time comparison}\label{fig:time}
\end{center}
\end{figure}

\paragraph{Computational Time and Size}
Let us examine the computational time for training the network. Fig.~\ref{fig:time} shows the comparison of the training time with LSTM-TD3 with $l=1$ serving as a benchmark.
Due to the added operations in the code, LSTM-TD3$_{1ha1hc}$ and LSTM-TD3$_{1ha2hc}$ exhibit a larger iteration time. The code is not optimized and there remains room for improvement in computational efficiency. 
As shown in the figure, H-TD3 provides a much shorter time per iteration and is less affected by the length $l$. This is because it avoids repeating the trajectory in critic networks.

\begin{figure*}[ht]
    \centering
    \includegraphics[width = \textwidth]{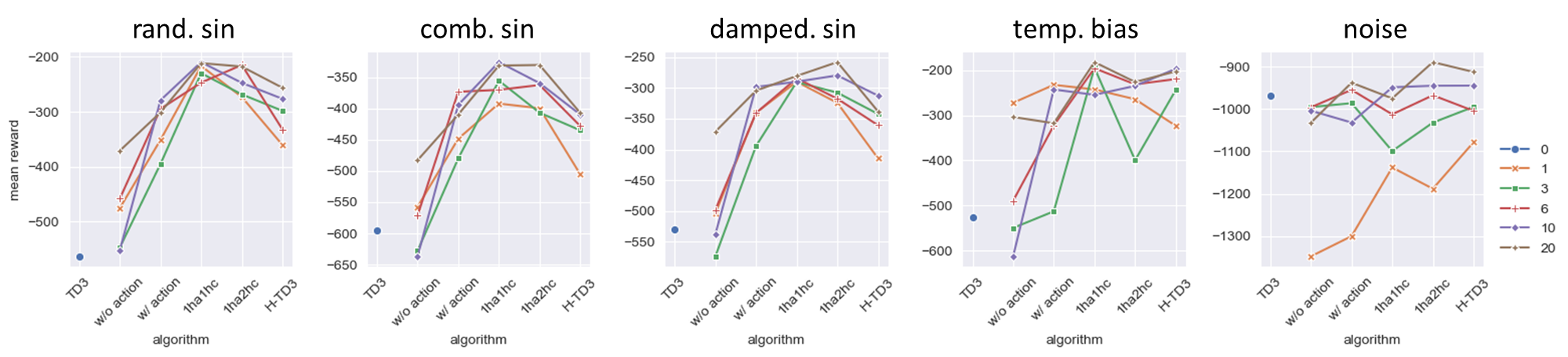}
    \caption{Generalisability}
    \label{fig:gene}
\end{figure*}

\paragraph{Generalisability}
So far, the networks have been evaluated in the same environment in which they were trained. However, in practice, the agent may experience different types of disturbances from the trained environment. For this reason, we tested how the networks trained in the “random sinusoidal wave” environment perform in “bias” and “noise” cases and also new environments: “combinational sinusoidal wave” and “damped sinusoidal wave”. In the “combinational sinusoidal wave”, two randomly configured sinusoidal waves with the same definition of “random sinusoidal wave” simultaneously appear, i.e., disturbance $= A_1\sin{2\pi t/T_1}+A_2\sin{2\pi t/T_2}$. On the other hand, the “damped sinusoidal wave” adds damped sinusoidal waves defined by $e^{-(t-t_0)/T} A\sin{2\pi (t-t_0)/T}$, on all $3$ observation elements. The time steps $t_0$ at which the damped sinusoidal waves begin are randomly selected, and each disturbance continues until the end of the episode. The “bias” cases with the amplitude $1.0$ and the “noise” cases with $\sigma=1.0$ are also selected in this experiment. The normalized total reward of the episode after 1,000 trials is illustrated in Fig.~\ref{fig:gene}. The “comb. sin” and “damped sin” represent “combinational sinusoidal wave” and “damped sinusoidal wave” respectively. On the first column, the result of the “random sinusoidal wave” is shown as a reference. The trained network performed well except in the “noise” condition. The dynamic model of disturbance exists in “combinational sinusoidal wave”, “damped sinusoidal wave” and “bias” cases. As discussed in Section~\ref{keyidea}, the required behavior is different between the “noise” case and the other three cases. This result reinforces the hypothesis that the network trained in “random sinusoidal wave” utilizes the dynamic model of the disturbance and the system dynamic model and is compatible with the environments that have temporally correlated disturbance.

\section{Engineering Case Study:
Differential-Drive Rover Motion Regulation under Hidden Wheel Slip}\label{engineering2}

To investigate whether the architectural insights identified in the controlled benchmark generalize to a physically motivated robotic control problem, we consider a simulation-based motion-regulation task for a differential-drive rover operating under hidden asymmetric wheel slip. Unlike the abstract disturbances studied in Section~\ref{discussion}, wheel slip arises naturally from terrain interaction and affects the rover through partially observable dynamics, providing a physically motivated test of the proposed recurrent architecture.

\subsection{Engineering motivation}
Planetary rovers operating on deformable terrain are susceptible to wheel slip. On loose lunar regolith, slip can become severe~\cite{coloma2022enhancing}, degrading rover mobility and, in extreme cases, leading to immobilization or entrapment that may compromise the mission.

Wheel slip originates from complex wheel--terrain interactions that are difficult to model accurately. Rather than reproducing the full terramechanical process, this work focuses on one characteristic that is particularly relevant to motion regulation: an evolving slip process that alters the relationship between commanded and realized rover motion.

During operation, the controller observes only the resulting vehicle motion through onboard proprioceptive sensing, such as inertial measurements and wheel odometry. The underlying slip dynamics, however, are not directly available to the controller. Consequently, similar observed vehicle motions may arise from different underlying slip dynamics, making the current observation alone insufficient for selecting an appropriate control action. The controller must therefore use the available action--observation history to infer decision-relevant information about the hidden slip dynamics.

Wheel slip arises from structured wheel--terrain interactions and therefore does not generally behave as an independent random disturbance at each control step \cite{ishigami2007terramechanics}. Motivated by this physical structure, the slip process in this study is modeled as a temporally correlated latent disturbance. Consequently, previous actions and their resulting observations provide additional information about the evolution of the hidden slip dynamics that is unavailable from the instantaneous observation alone.

This characteristic makes rover motion regulation under evolving hidden slip an appropriate engineering-inspired case study for evaluating whether incorporating temporal action--observation histories improves decision making under evolving hidden dynamics.

\subsection{System model}
The rover is modeled as a differential-drive vehicle. Its body linear velocity and yaw rate are determined by the realized left and right wheel velocities according to the standard differential-drive kinematics:
\begin{align}
v &= \frac{v_L+v_R}{2},\\
\omega &= \frac{v_R-v_L}{b},
\end{align}
where $v_L$ and $v_R$ are realized wheel velocities and the wheelbase is $b=0.5$ m. 

To represent mobility degradation, each wheel is affected by an independent hidden slip variable that attenuates its nominal velocity. Rather than explicitly modeling wheel--terrain interaction, the slip variables provide an engineering abstraction of traction loss.

\begin{align}
v_L &= (1-\lambda_L)\,\bar v_L,\\
v_R &= (1-\lambda_R)\,\bar v_R,
\end{align}
where $\bar{v}_L$ and $\bar{v}_R$ denote the nominal wheel velocities in the absence of slip, while $\lambda_L,\lambda_R \in [0,1]$ denote the hidden slip variables. A value of $\lambda_i=0$ corresponds to nominal traction, whereas $\lambda_i=1$ represents complete attenuation of the corresponding wheel velocity. In this study, the slip variables are restricted to $\lambda_i \in [0,\lambda_{\max}]$, with $\lambda_{\max}=0.5$, to focus on motion regulation under partial traction loss rather than complete wheel immobilization.

\subsection{RL setup}
The rover motion regulation task is formulated as an RL problem. At each control step, the agent selects velocity commands based on the observable tracking errors, while the underlying wheel slip remains unobservable.

\begin{itemize}
    \item \textbf{Action:}
    \[
    \mathbf{a}_t=
    \begin{bmatrix}
    v_{\rm cmd}\\
    \omega_{\rm cmd}
    \end{bmatrix},
    \]
    where the corresponding nominal wheel velocities are
    \begin{align}
    \bar v_L &= v_{\rm cmd}-\frac{b}{2}\omega_{\rm cmd},\\
    \bar v_R &= v_{\rm cmd}+\frac{b}{2}\omega_{\rm cmd}.
    \end{align}

    \item \textbf{Observation:}
    \[
    \mathbf{o}_t=
    \begin{bmatrix}
    e_v\\
    e_\omega
    \end{bmatrix}
    =
    \begin{bmatrix}
    v-v_{\rm target}\\
    \omega-\omega_{\rm target}
    \end{bmatrix}.
    \]

    \item \textbf{Reward:}
    \[
    r_t=-(e_v^2+e_\omega^2).
    \]
\end{itemize}

The wheel-slip variables $\lambda_L$ and $\lambda_R$ are omitted from the observation, while the observed tracking errors $e_v$ and $e_\omega$ reflect both the commanded motion and the attenuation caused by slip. Because the realized linear and angular velocities are jointly determined by the applied action and the current left- and right-wheel slip, the observation alone does not identify the slip state. By combining the current observation with the recent history of actions and observations, the policy can infer the effective wheel-slip state and capture its temporal evolution. This context supports both estimation of the current slip and selection of the next action based on its recent evolution, making the control task partially observable.

Each episode starts from
$v_0=0$ m/s,
$\omega_0=0$ rad/s,
and aims to regulate the rover to
$v_{\rm target}=0.5$ m/s,
$\omega_{\rm target}=0$ rad/s.
Each episode comprises 200 control steps.

\subsection{Experimental evaluation}

The engineering case study evaluates whether the conclusions obtained from the benchmark environments extend to rover motion regulation under hidden wheel slip. Specifically, Exp1 examines whether temporal action--observation history and the recurrent architecture remain important when the hidden disturbance is introduced through the rover dynamics. Exp2 investigates whether policies trained in the abstract disturbance environments generalize to the rover environment without additional fine-tuning. 
\subsubsection{Experimental setting}
\paragraph{Exp1: Validation under Hidden Wheel Slip}
The rover is trained and evaluated under hidden wheel slip. The left and right wheel slip variables evolve independently according to clipped first-order autoregressive (AR(1)) processes,
\begin{align}
\lambda_{i,t+1}
=
\operatorname{clip}
\left(
\rho\lambda_{i,t}
+
\epsilon_{i,t},
0,
\lambda_{\max}
\right),
\qquad
i\in\{L,R\},
\end{align}
where $\rho$ controls the temporal correlation, $\epsilon_{i,t}\sim\mathcal{N}(0,\sigma^2)$, and $\operatorname{clip}(\cdot)$ limits the slip to $[0,\lambda_{\max}]$. Unless otherwise stated, the nominal slip model uses $\rho=0.95$, $\sigma=0.05$, and $\lambda_{\max}=0.5$, with the initial slip sampled as $\lambda_{L/R,0}\sim \operatorname{Unif}(0,0.5)$.

The same nominal AR(1) slip model is used for training and evaluation, providing a baseline for validating the benchmark findings in the rover environment.

\paragraph{Exp2: Generalization from Abstract Disturbances}
Exp2 evaluates whether policies trained under an abstract disturbance model transfer to the rover environment without further training. During training, the physically motivated AR(1) slip process is replaced by a sinusoidal slip model,

\begin{align}
\lambda_{i,t}
=
\operatorname{clip}
\left(
A_i
\sin\left(
\frac{2\pi t}{T_i}
+\phi_i
\right)
+\lambda_0,
0,
\lambda_{\max}
\right),
\qquad
i\in\{L,R\},
\end{align}
where the amplitude $A_i$, period $T_i$, and phase $\phi_i$ are independently sampled for each wheel at the beginning of every episode, with
$A_i \sim \operatorname{Unif}(0.08, 0.22)$,
$T_i \sim \operatorname{Unif}\{40,\ldots,90\}$,
$\phi_i \sim \operatorname{Unif}(0, 2\pi)$,
$\lambda_0 = 0.12$, and
$\lambda_{\max} = 0.5$.
The resulting slip variables are applied to the rover using the same multiplicative slip model as in Exp1. The disturbance parameters were selected to approximately match the slip distribution produced by the physically motivated AR(1) model. During evaluation, the abstract disturbance is replaced by the AR(1) wheel-slip process, and the policy is tested without further training.
As an additional evaluation, the trained policies are also tested under modified AR(1) slip dynamics by changing the standard deviation of the zero-mean Gaussian distribution and the temporal correlation. Specifically, the high-slip-variance setting increases the standard deviation to $\sigma=0.10$, while the fast- and slow-slip settings modify the temporal correlation to $\rho=0.90$ and $\rho=0.98$, respectively. All remaining parameters are unchanged.

\paragraph{Training protocol}
All methods use the same training protocol and hyperparameters as in the benchmark experiments unless otherwise stated. Each policy is trained using five random seeds and evaluated over $20$ episodes per seed, giving $100$ evaluation episodes per condition. The reported mean and standard deviation are computed across the five seed-level mean returns.

\subsection{Results and Discussion}
This section evaluates whether the findings from the benchmark environments extend to rover motion regulation under hidden wheel slip. Exp1 first evaluates the proposed architectures when trained and tested under the physically motivated AR(1) wheel-slip model. Exp2 then investigates whether policies trained under the proposed abstract disturbance model generalize to the simulated wheel-slip environment and evaluates their robustness under modified slip dynamics.

\subsubsection{Exp1: Validation under Hidden Wheel Slip}

Table~\ref{tab:results} summarizes the performance of all architectures when trained and evaluated under the physically motivated AR(1) wheel-slip model. The architectural trends observed in the benchmark environments are largely preserved in the rover motion regulation task. The recurrent architectures outperform the feedforward TD3 baseline. Among the action-history-based recurrent methods, LSTM-TD3 with action, LSTM-TD3\textsubscript{1ha1hc}, and LSTM-TD3\textsubscript{1ha2hc} achieve comparable performance, with LSTM-TD3\textsubscript{1ha1hc} attaining the highest mean return.

Unlike the benchmark environments, where the hidden disturbance is introduced through additive observation disturbances, the rover disturbance arises from multiplicative wheel slip that modifies the realized control inputs. Despite this fundamentally different disturbance mechanism, the action-history-based recurrent methods retain their advantage. These results indicate that the benchmark findings extend beyond the abstract control problems considered in Section~\ref{discussion} to a physically motivated robotic control task.

\subsubsection{Exp2: Generalization from Abstract Disturbances}

Table~\ref{tab:results} further compares policies trained under different disturbance models when evaluated on the simulated AR(1) wheel-slip environment. Across all architectures, training with the proposed abstract disturbance consistently improves transfer performance compared with training without disturbances. Although the target wheel-slip dynamics are never observed during training, policies trained under the abstract disturbance achieve substantially higher returns than policies trained without disturbances. These results demonstrate that the proposed abstract disturbance provides an effective training surrogate for improving transfer to the simulated wheel-slip environment.

Among the recurrent architectures, LSTM-TD3\textsubscript{1ha1hc} exhibits particularly strong transfer capability. While LSTM-TD3 with action, LSTM-TD3\textsubscript{1ha1hc}, and LSTM-TD3\textsubscript{1ha2hc} achieve nearly identical performance when trained directly on the physically motivated AR(1) wheel-slip model, their transfer performance differs considerably. In particular, LSTM-TD3\textsubscript{1ha1hc} remains close to its in-distribution performance, whereas LSTM-TD3 with action experiences a substantially larger degradation after transfer. This indicates that LSTM-TD3\textsubscript{1ha1hc} provides stronger robustness to the previously unseen disturbance dynamics considered here, rather than only achieving high performance within the training distribution.

\begin{figure*}[t]
    \centering
    \includegraphics[width=\textwidth]{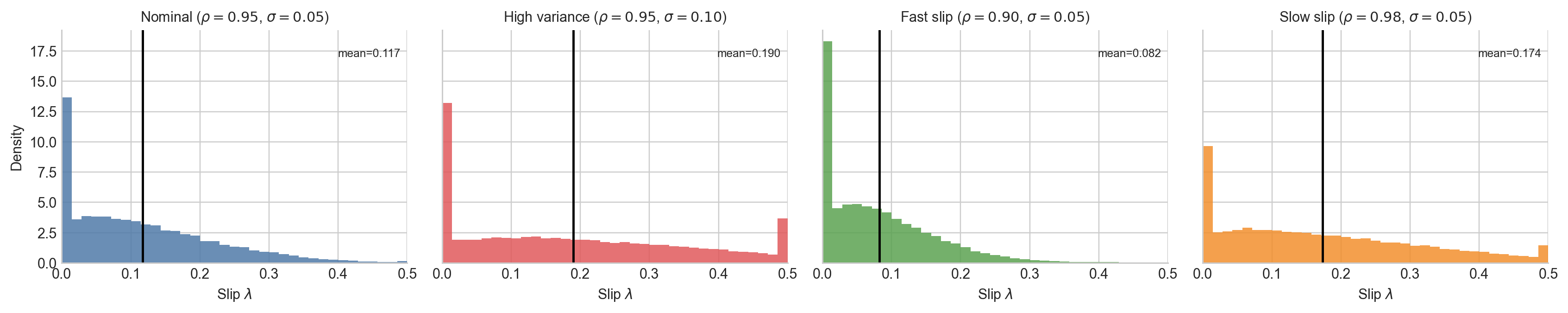}
    \caption{Marginal slip distributions under the Nominal ($\rho=0.95$, $\sigma=0.05$), High variance ($\rho=0.95$, $\sigma=0.10$), Fast slip ($\rho=0.90$, $\sigma=0.05$), and Slow slip ($\rho=0.98$, $\sigma=0.05$) conditions. Vertical lines indicate the mean slip magnitude, with $\mu_{\lambda}=0.117$, $0.190$, $0.082$, and $0.174$, respectively.}
    \label{fig:slip_distributions}
\end{figure*}

To further evaluate robustness, the policies trained under the proposed abstract disturbance are evaluated without additional training under modified wheel-slip conditions. Specifically, the nominal AR(1) wheel-slip model is replaced by variants with increased stochastic variability (High variance) and modified persistence parameters. As summarized in Table~\ref{tab:results}, performance degrades most strongly under the High variance condition. As shown in Fig.~\ref{fig:slip_distributions}, this condition produces a broader slip distribution in which stochastic variation becomes dominant over the temporally correlated evolution, making transfer substantially more difficult. This behavior is consistent with the pendulum benchmark, where uncorrelated noise was more difficult under in-domain training and policies trained under sinusoidal disturbances failed to generalize to the noise condition, as discussed in Section~\ref{discussion}.

The persistence parameter also affects the marginal slip distribution. The Fast slip condition places substantially more probability mass near zero and has a lower mean slip magnitude than the Nominal and Slow slip conditions, which likely contributes to its better performance. In contrast, the Slow slip condition produces more persistent and sustained attenuation together with a higher mean slip magnitude, resulting in a more challenging control problem.

\begin{table*}[t]
\centering
\caption{Performance on the simulation-based wheel-slip task after training under different disturbance conditions. Values are reported as the mean $\pm$ standard deviation across five seed-level average returns, with each seed-level average computed over 20 evaluation episodes. The robustness columns report evaluation of the abstract-disturbance-trained policies under modified wheel-slip dynamics. Bold values indicate the highest mean return in each column.}
\label{tab:results}

\resizebox{\textwidth}{!}{%
\begin{tabular}{lccc|ccc}
\toprule
&
\multicolumn{3}{c|}{Training condition}
&
\multicolumn{3}{c}{Robustness evaluation} \\
\cmidrule(lr){2-4}
\cmidrule(lr){5-7}
Method &
AR(1) wheel slip &
Abstract &
None &
High variance &
Fast slip &
Slow slip \\
\midrule

TD3
& $-2.92 \pm 0.07$
& $-2.94 \pm 0.09$
& $-5.21 \pm 0.19$
& $-7.75 \pm 0.17$
& $-1.73 \pm 0.17$
& $-5.33 \pm 0.14$ \\

LSTM-TD3 w/ action
& $-1.38 \pm 0.08$
& $-2.40 \pm 0.29$
& $-5.26 \pm 0.14$
& $-7.38 \pm 0.97$
& $-2.17 \pm 0.24$
& $-2.78 \pm 0.38$ \\

LSTM-TD3 w/o action
& $-1.78 \pm 0.07$
& $-2.55 \pm 0.24$
& $-5.29 \pm 0.16$
& $-6.90 \pm 0.18$
& $-2.11 \pm 0.30$
& $-3.76 \pm 0.19$ \\

LSTM-TD3\textsubscript{1ha1hc}
& $\mathbf{-1.36 \pm 0.04}$
& $\mathbf{-1.70 \pm 0.06}$
& $-5.20 \pm 0.14$
& $\mathbf{-5.95 \pm 0.22}$
& $\mathbf{-1.48 \pm 0.10}$
& $\mathbf{-2.05 \pm 0.06}$ \\

LSTM-TD3\textsubscript{1ha2hc}
& $-1.39 \pm 0.05$
& $-2.08 \pm 0.16$
& $-5.19 \pm 0.04$
& $-6.09 \pm 0.24$
& $-1.89 \pm 0.12$
& $-2.37 \pm 0.10$ \\

H-TD3
& $-1.92 \pm 0.07$
& $-2.28 \pm 0.17$
& $\mathbf{-5.13 \pm 0.23}$
& $-7.93 \pm 1.26$
& $-1.91 \pm 0.20$
& $-3.17 \pm 0.31$ \\

\bottomrule
\end{tabular}%
}
\end{table*}

\section{Conclusion}\label{conclusion}

In this paper, we investigated recurrent extensions of TD3 for continuous-control problems under partial observability and time-varying disturbances. When the disturbance evolves over time, the policy must infer the resulting hidden dynamics from interaction history and adapt its decisions accordingly. In this context, observation history alone may be insufficient to infer the latent dynamics when the observed system response also depends on the previously applied actions. More generally, decision-relevant internal representations in a POMDP can be constructed from both observation and action histories, since together they provide information about the evolution of the latent state. This principle also applies to end-to-end model-free RL, where the internal representation is learned directly from interaction data. We showed that incorporating action history can improve performance under evolving disturbances without explicitly separating the policy into state-estimation and control modules.

We also examined how recurrent architectures should process past and current information within the actor and critic networks. The results showed that processing past and current action--observation information within a unified temporal sequence improves performance compared with processing them through separate branches. We also proposed a new method called H-TD3 that reuses recurrent states generated by the actor to initialize the critic, thereby reducing duplicated sequence processing between the two networks. The simulation-based engineering case study further showed that the advantage of action-history-based recurrent architectures extends to differential-drive rover motion regulation under hidden asymmetric wheel slip, where the disturbance enters through multiplicative changes in the realized wheel velocities. Moreover, policies trained under abstract temporally structured disturbances transferred more effectively to previously unseen wheel-slip dynamics than policies trained without disturbances, with LSTM-TD3\textsubscript{1ha1hc} showing particularly strong transfer performance across the evaluated slip conditions.

For future work, the generalization analysis should be extended to higher-fidelity wheel--terrain models and physical rover experiments. The results also showed that recurrent architectures perform differently depending on the temporal characteristics of the disturbance. Future work should therefore investigate architectures that remain effective across both temporally structured disturbances and stochastic noise, while validating their performance under more realistic and diverse engineering conditions.

\section*{Acknowledgments}

The original methodological study was supported by Cranfield University and QuadSAT. The engineering case study was developed as part of the NeverGiveUp project. This research was funded in whole, or in part, by the Luxembourg National Research Fund (FNR), grant reference C24/IS/18990533/NeverGiveUp. For the purpose of open access, and in fulfilment of the obligations arising from the grant agreement, the author has applied a Creative Commons Attribution 4.0 International (CC BY 4.0) licence to any Author Accepted Manuscript version arising from this submission.

\appendix
\section{Appendix}

\subsection{Network Structure}
The network structures used are summarized in Table \ref{tab:structure}.
\begin{table*}[ht]
\centering
\caption{Structure}
\label{tab:structure}

\begin{tabularx}{\textwidth}{l X X}
\hline \hline
Algorithm & Actor network & Critic network\\
\hline
LSTM-TD3 & \makecell[l]{Lin(obs dim+act dim, 128), Lin(obs dim, 128)\\
ReLU, ReLU\\
LSTM(128,128), N/A\\
Lin(256,128)\\
ReLU\\
Lin(128,1)\\
tanh} & \makecell[l]{Lin(obs dim+act dim, 128), Lin(obs dim\\ \hspace{4.2cm} +act dim, 128)\\
ReLU, ReLU\\
LSTM(128,128), N/A\\
Lin(256,128)\\
ReLU\\
Lin(128,1)} \\
\hline
LSTM-TD3$_{1ha1hc}$ &  \makecell[l]{Lin(obs dim+act dim, 128)\\ReLU\\LSTM(128,128)\\Lin(128,128)\\ReLU\\Lin(128,1)\\ tanh}  &  \makecell[l]{Lin(obs dim+act dim, 128)\\ReLU\\LSTM(128,128)\\Lin(128,128)\\ReLU\\Lin(128,1)} \\
\hline
LSTM-TD3$_{1ha2hc}$ & \makecell[l]{Lin(obs dim+act dim, 128)\\ReLU\\LSTM(128,128)\\Lin(128,128)\\ReLU\\Lin(128,1)\\ tanh}  &  \makecell[l]{Lin(obs dim+act dim, 128), Lin(act dim, 128) \\ReLU, ReLU\\LSTM(128,128), N/A\\Lin(256,128)\\ReLU\\Lin(128,1)} \\
\hline
H-TD3 &  \makecell[l]{Lin(obs dim+act dim, 128)\\ReLU\\LSTM(128,128)\\Lin(128,128)\\ReLU\\Lin(128,1)\\ tanh}  &\makecell[l]{Lin(obs dim+act dim, 128)\\ReLU\\LSTM(128,128)\\Lin(128,128)\\ReLU\\Lin(128,1)} \\
\hline
TD3 & \makecell[l]{Lin(obs dim, 128)\\ReLU\\Lin(128,128)\\ReLU\\Lin(128,1)\\ tanh} & \makecell[l]{Lin(obs dim+act dim, 128)\\ReLU\\Lin(128,128)\\ReLU\\Lin(128,1)}\\
\hline\hline
\multicolumn{3}{l}{Lin: linear layer}
\end{tabularx}
\end{table*}

\subsection{The Number of Parameters}
The numbers of parameters are summarized in Table \ref{tab:parameter}.

\begin{table}[H]
\begin{center}
\caption{Number of parameters (obs dim = 3, act dim = 1)} \label{tab:parameter}
\begin{tabular}{cc}
\hline \hline
Algorithm & The number of parameters (actor, critic) \\
\hline
LSTM-TD3 &166273, 166401 \\
\hline
LSTM-TD3$_{1ha1hc}$ & 149377, 149377\\
\hline
LSTM-TD3$_{1ha2hc}$ &149377, 166017\\
\hline
H-TD3 &149377, 149377 \\
\hline
TD3 & 17153, 17281 \\
\hline\hline
\end{tabular}
\end{center}
\end{table}

\bibliographystyle{unsrt}  
\bibliography{references}

\end{document}